\documentclass{article} 
\usepackage{iclr2023_conference, times}
\pdfoutput=1

\usepackage{amsmath,amsfonts,bm}

\def\eqref#1{equation~\ref{#1}}

\def\1{\bm{1}}

\DeclareMathAlphabet{\mathsfit}{\encodingdefault}{\sfdefault}{m}{sl}
\SetMathAlphabet{\mathsfit}{bold}{\encodingdefault}{\sfdefault}{bx}{n}

\newcommand{\softmax}{\mathrm{softmax}}

\usepackage{graphicx}
\usepackage{hyperref}
\usepackage{url}
\usepackage{tabularx}
\usepackage{booktabs}
\usepackage{wrapfig}
\usepackage{float}
\usepackage{comment} 

\usepackage{multirow}

\title{Distil-Whisper: Robust Knowledge \\Distillation via Large-Scale \\Pseudo Labelling}

\author{Sanchit Gandhi, Patrick von Platen \& Alexander M. Rush  \\
Hugging Face \\
\texttt{\{sanchit, patrick, sasha\}@huggingface.co} \\
}

\iclrfinalcopy 

\renewcommand{\headrulewidth}{0pt}
\renewcommand{\footrulewidth}{0pt}

\begin{document}

\maketitle

\begin{abstract}

As the size of pre-trained speech recognition models increases, running these large models in low-latency or resource-constrained environments becomes challenging. In this work, we leverage pseudo-labelling to assemble a large-scale open-source dataset which we use to distill the Whisper model into a smaller variant, called Distil-Whisper. Using a simple word error rate (WER) heuristic, we select only the highest quality pseudo-labels for training. The distilled model is 5.8 times faster with 51\% fewer parameters, while performing to within 1\% WER on out-of-distribution test data in a zero-shot transfer setting. Distil-Whisper maintains the robustness of the Whisper model to difficult acoustic conditions, while being less prone to hallucination errors on long-form audio. Distil-Whisper is designed to be paired with Whisper for speculative decoding, yielding a 2 times speed-up while mathematically ensuring the same outputs as the original model. To facilitate further research in this domain, we make our training code, inference code and models publicly accessible.

\end{abstract}

\section{Introduction}
In recent years, Automatic Speech Recognition (ASR) systems have surpassed human-level accuracy in many academic benchmarks \citep{amodei16_deepspeech, baevski20_wav2vec2, zhang22_pushing}, enabling a wide range of applications from transcription services to voice assistants \citep{aksenova21_asrbenchmarks}. Whisper \citep{radford22_whisper}, a 1.5 billion parameter sequence-to-sequence (Seq2Seq) transformer model \citep{vaswani17_attention} pre-trained on 680,000 hours of weakly supervised speech recognition data, demonstrates a strong ability to generalise to many different datasets and domains \citep{gandhi22_esb}. However, the ever-increasing size of pre-trained ASR models poses challenges when deploying these systems in low-latency settings or resource-constrained hardware \citep{he18_conformer, zhang22_bigssl}.

Recent efforts in natural language processing (NLP) have demonstrated promising advancements in compressing transformer-based models. Knowledge distillation (KD) has successfully been applied to reduce the size of models such as BERT \citep{devlin19_bert}, without a significant performance loss on non-generative classification tasks \citep{sanh19_distilbert, jiao20_tinybert, sun20_mobilebert}. Inspired by machine translation methods, pseudo-labelling (PL) approaches \citep{kim16_sequence} have also been explored for Seq2Seq summarisation \citep{schleifer20_distilbart}, demonstrating the potential for substantial compression of Seq2Seq models on generative tasks. In the audio domain, KD has shown promising results for audio classification \citep{peng21_shrinking, chang22_distilhubert}. However, similar results have not yet been achieved for the more difficult task of speech recognition.

In this paper, we apply distillation to the Whisper model in the context of Seq2Seq ASR. We address the challenge of maintaining robustness to different acoustic conditions through our construction of a large-scale open-source dataset covering 10 distinct domains. By pseudo-labelling the data, we ensure consistent transcription formatting across the dataset and provide sequence-level distillation signal. We propose a simple word error rate (WER) based heuristic for filtering the pseudo-labelled data and demonstrate that it is an effective method for ensuring good downstream performance of the distilled model.

We demonstrate that Distil-Whisper maintains the robustness of Whisper to different audio domains and noisy acoustic conditions. We measure this by evaluating the distilled models on four out-of-distribution test sets spanning multiple audio domains. The best model performs to within 1\% WER of original Whisper checkpoint, while being 5.8 times faster with 51\% fewer parameters. On long-form evaluation, the distilled model outperforms Whisper by 0.1\% WER. We show that this performance gain is due to a lower propensity to hallucinate than the original Whisper model. 

By sharing the same encoder weights as Whisper, Distil-Whisper can be used efficiently as an assistant model to Whisper for speculative decoding \citep{leviathan23_speculative}, for which we achieve a 2 times improvement in inference speed with only an 8\% increase to parameter count. Speculative decoding algorithmically ensures that predictions of the main model are unchanged, meaning it can be used as a drop-in replacement for existing speech recognition pipelines using Whisper.

Our work suggests that large-scale pseudo-labelling of speech data has been under-explored and that it provides a promising technique for KD. To serve as a basis for further research on distillation for speech recognition, we release training code, inference code and models under \url{https://github.com/huggingface/distil-whisper}.

\section{Related Work}

In the NLP domain, model distillation has demonstrated substantial promise in reducing model size and computational requirements with minimal degradation to performance. \cite{sanh19_distilbert} use a weighted average of the KD loss and the traditional cross entropy data loss to train DistilBERT, a 6 layer distilled version of BERT \citep{devlin19_bert}, that achieves a 40\% decrease in model size, a 60\% increase in speed, and a 97\% preservation of language under-standing capabilities on the GLUE benchmark \citep{wang19_glue}. \cite{schleifer20_distilbart} extend the DistilBERT methodology the Seq2Seq setting, by initialising the student decoder from maximally spaced layers of the teacher decoder and incorporating intermediate hidden-states into the KD loss function. The resulting model, DistilBART, outperforms the original model on the XSUM and CNN/Daily Mail datasets \citep{narayan18_xsum, see17_cnn}, with 37\% model compression and a 48\% increase in speed. \cite{du23_robustness} demonstrate that while distilled models perform well on in-distribution (ID) evaluation data, they perform significantly worse than their pre-trained counterparts on out-of-distribution (OOD) test sets. By training on a diverse, large-scale pseudo-labelled dataset, we preserve the robustness to different acoustic conditions, demonstrated by an ability to generalise to OOD test data.

KD has also been applied to the ASR task, albeit with a focus on encoder-only models. \cite{peng21_shrinking} apply KD to the Wav2Vec 2.0 model \citep{baevski20_wav2vec2}, achieving 79\% model compression and 59\% increase in speed. However, these gains come at the expense of a 6.9\% 
increase to WER on the LibriSpeech corpus \citep{panayotov15_librispeech}. \cite{chang22_distilhubert} apply a similar method to the HuBERT model \citep{hsu21_hubert}, and too report a 7.0\% 
WER increase. \cite{pang18_compression} attempt to distill LAS \cite{chan16_las}, an early Seq2Seq ASR model, but find their best distilled model performs 2.2\% WER worse than its larger counterpart. This paper focuses on KD of Seq2Seq models, with substantial model compression but also preserving WER performance on OOD test data.

Previous studies involving distilling the Whisper model have predominantly been centered around reducing model size and memory footprint. \cite{shao23_whisper_kdq} applied KD in combination with Quantisation Aware Training (QAT) \citep{jacob17_quantisation}, demonstrating that significant parameter reduction is possible with only marginal performance decrement. However, the student model is trained and tested on a small corpus of ID data, giving no measure of its ability to generalise to OOD data, and thus its robustness to different acoustic conditions \citep{geirhos20_shortcut, radford22_whisper}. Furthermore, this work did not consider optimising the model for latency. This paper seeks to distill the Whisper model to achieve significant model compression, jointly with latency improvements and WER performance on OOD test data. We also evaluate the distilled models' robustness to noisy audio conditions.

\section{Background}
Whisper \citep{radford22_whisper} is a sequence-to-sequence (Seq2Seq) transformer model \citep{vaswani17_attention} pre-trained on 680,000 hours of noisy speech recognition data web-scraped from the internet. When scaled to this quantity of data, Whisper yields competitive results with fully supervised systems, but in a \textit{zero-shot} setting without the need for any fine-tuning.

Whisper is composed of a transformer-based encoder (Enc) and decoder (Dec). Assume we have an input speech signal comprised of $T$ feature vectors $\boldsymbol{X}_{1:T} = \{\boldsymbol{x}_1, \ldots, \boldsymbol{x}_T \}$ and a target transcription $\boldsymbol{y}_{1:N} = \{ y_1, \ldots, y_N \}$ of $N$ tokens in the standard speech recognition setting. The encoder $\mathcal{F}_{Enc}$ is trained to map $\boldsymbol{X}_{1:T}$ to a sequence of hidden-state vectors $\boldsymbol{H}_{1:M}$:

\begin{align}
    \mathcal{F}_{Enc}: \boldsymbol{X}_{1:T} \rightarrow \boldsymbol{H}_{1:M}
\end{align}

The sequence length of the hidden-states $M$ is typically half than that of the input speech feature sequence $T$ by action of the convolutional layers in the encoder stem that downsample the input.

The decoder auto-regressively predicts a probability distribution for the next token $y_{i}$, conditional on all previous tokens $\boldsymbol{y}_{<i}$ and the encoder hidden-states $\boldsymbol{H}_{1:M}$:

\begin{align}
    P\left( y_{i} | \boldsymbol{y}_{<i}, \boldsymbol{H}_{1:M} \right)
\end{align}

To train the Whisper model, we assume a dataset where each example $\left( \boldsymbol{X}_{1:T}, \boldsymbol{y}_{1:N} \right)$ is an (audio, text) pair. The model is trained using the standard cross-entropy (CE) loss, where the model is trained to predict an instance class by maximising the estimated probability of the target class labels:

\begin{align}
    \mathcal{L}_{CE} = - \sum_{i=1}^{N} P\left( y_{i} | \boldsymbol{y}_{<i}, \boldsymbol{H}_{1:M} \right) \label{eq: cross-entropy}
\end{align}

There are five variants of the Whisper model summarised in Table~\ref{tab: whisper checkpoints}. The models share the same Seq2Seq architecture but have different dimensionality. For all model sizes, the encoder and decoder have the same width, heads and number of layers in the transformer blocks. The first version of the Whisper paper introduced a large-v1 checkpoint, which was subsequently re-trained with regularisation more training epochs to give an improved large-v2 version \citep{radford22_whisper}. As both models share the same dimensions, we present results for the large-v2 model in this paper.

\begin{table}[t]
\caption{Dimensionality details of the pre-trained Whisper checkpoints.}
\begin{center}
\label{tab: whisper checkpoints}
\begin{tabular}{@{}lrrrr@{}}
\toprule
\textbf{Model}     & \multicolumn{1}{l}{\textbf{Layers}} & \multicolumn{1}{l}{\textbf{Width}} & \multicolumn{1}{l}{\textbf{Heads}} & \multicolumn{1}{l}{\textbf{Parameters / M}} \\ \midrule
\href{https://huggingface.co/openai/whisper-tiny.en}{tiny.en}     & 4                          & 384                       & 6                         & 39                                 \\
\href{https://huggingface.co/openai/whisper-base.en}{base.en}     & 6                          & 512                       & 8                         & 74                                 \\
\href{https://huggingface.co/openai/whisper-small.en}{small.en}    & 12                         & 768                       & 12                        & 244                                \\
\href{https://huggingface.co/openai/whisper-medium.en}{medium.en}   & 24                         & 1024                      & 16                        & 769                                \\
\href{https://huggingface.co/openai/whisper-large-v2}{large-v2} & 32                         & 1280                      & 20                        & 1550                               \\ \bottomrule
\end{tabular}
\end{center}
\end{table}

\section{Robust Knowledge-Distillation}
\subsection{Knowledge Distillation} \label{sec: knowledge distillation}

Knowledge distillation (KD) \citep{hinton15_knowledge} is a compression technique in which a smaller student model is trained to reproduce the behaviour of a larger teacher one. Compared to minimising the CE loss between the student model's predictions and the training labels, KD allows the student model to learn from the full predictive distribution of possible next tokens in a given context, rather than just maximising the probability of the next target token in the training data.

\paragraph{Shrink and Fine-Tune} The most basic distillation method involves shrinking the teacher model to a smaller student size, and training the student on the CE objective in Equation \ref{eq: cross-entropy}. Following \cite{schleifer20_distilbart}, we perform layer-based compression by initialising the student model by copying the weights from maximally spaced layers of the teacher model. For example, when initialising a 2-layer student model from a 32-layer teacher model, we copy the 1\textsuperscript{st} and 32\textsuperscript{nd} layers from the teacher to the student. Given the simplicity and effectiveness of this strategy in the Seq2Seq summarisation setting \citep{schleifer20_distilbart, li22_dqbart}, we use it for all distillation methods.

\paragraph{Pseudo Labelling} In the pseudo-label setting \citep{kim16_sequence}, we replace the ground truth text transcription $\boldsymbol{y}_{1:N}$ with the teacher's generation $\hat{\boldsymbol{y}}_{1:N'}$ for the corresponding input audio $\boldsymbol{X}_{1:T}$:

\begin{align}
    \mathcal{L}_{PL} = - \sum_{i=1}^{N'} P\left( y_{i} | 
    \hat{\boldsymbol{y}}_{<i}, \boldsymbol{H}_{1:M} \right)
\end{align}

This form of distillation can be viewed as ``sequence-level" KD, where knowledge is transferred from the teacher model to the student model across a sequence of generated pseudo-labels \citep{kim16_sequence}.

\paragraph{Kullback-Leibler Divergence} In the KL Divergence \citep{kullback51_divergence} setting, the full probability distribution of the student model $P_{i}$ is trained to match the full distribution of the teacher model $Q_{i}$ by minimising the KL divergence over the entire set of next possible tokens at position $i$:

\begin{align}
    \mathcal{L}_{KL} &= \sum_{i=1}^{N} KL \left(Q_{i}, P_{i} \right) 
\end{align}

This can be viewed as ``word-level" KD, where knowledge is transferred from the teacher model to the student model via the logits over the possible tokens \citep{kim16_sequence}. The KL Divergence is attractive since it provides information over all classes and has less variance in gradients than the CE loss \citep{hinton15_knowledge}.

\paragraph{Objective} The final KD training objective is a weighted sum of the KL and PL terms:

\begin{align}
    \mathcal{L}_{KD} = \alpha_{KL} \mathcal{L}_{KL} + \alpha_{PL} \mathcal{L}_{PL}
\end{align}

where $\alpha_{KL}$ and $\alpha_{PL}$ are scalar weights for the KL and loss terms respectively. Following \citep{schleifer20_distilbart}, we set $\alpha_{KL} = 0.8$ and $\alpha_{PL} = 1.0$.

\subsection{Pseudo-Label Selection: WER Threshold}

The pseudo-labels generated by the Whisper model are subject to transcription errors and hallucinations \citep{bain23_whisperx, zhang23_usm}. To ensure we only train on accurate pseudo-labels, we propose a simple heuristic to filter our pseudo-labelled training data. For each training sample, we normalise the ground truth labels $\boldsymbol{y}_{1:N}$ and the Whisper generated pseudo-labels $\hat{\boldsymbol{y}}_{1:N'}$ using the Whisper English normaliser \citep{radford22_whisper}. We compute the word error rate (WER) between the normalised ground truth and normalised psuedo-labels, and discard any samples that exceed a given WER threshold $\lambda$:

\begin{equation}
    \text{WER}\left(\text{norm}\left(\boldsymbol{y}_{1:N}\right), \text{norm}\left(\hat{\boldsymbol{y}}_{1:N'}\right)\right) > \lambda
\end{equation}

We tune the value of $\lambda$ on our validation sets, and demonstrate in Section \ref{sec: wer threshold analysis} that this simple filtering method improves transcription quality and downstream model performance.

\section{Chunked Long-Form Transcription}

Whisper models have a fixed receptive field corresponding to 30-seconds of input audio and cannot process longer audio inputs at once. Most academic datasets comprise of short utterances less than 30-seconds in duration, and so this is not a problem. However, real-world applications such as meeting transcriptions typically require transcribing long audio files of many minutes or hours.

The original Whisper paper presents a long-form transcription algorithm that sequentially transcribes 30-second segments of audio and shifts the sliding window according to the timestamps predicted by the model. This auto-regressive algorithm requires both beam-search and temperature fallback to ensure accurate long-form transcription \citep{radford22_whisper}.

We use an alternative strategy, first proposed by \cite{patry22_chunking}, in which the long audio file is chunked into smaller segments with a small overlap between adjacent segments. The model is run over each chunk and the inferred text is joined at the strides by finding the longest common sequence between overlaps. This stride enables accurate transcription across chunks without having to transcribe them sequentially. We observe that this algorithm only requires greedy decoding to reliably transcribe long audio files. Furthermore, this algorithm is semi auto-regressive in the sense that the chunks can be transcribed in any order, provided adjacent chunks are subsequently joined correctly at their boundaries. This allows chunks to be transcribed in parallel through batching. In practice, this yields up to 9 times improvements in inference speed compared to sequential transcription over long audio files. In this work, use the chunked long-form transcription algorithm when evaluating both the Whisper and Distil-Whisper models. 

\section{Speculative Decoding}

Speculative decoding (SD) \citep{leviathan23_speculative} is a method for accelerating the inference of auto-regressive transformer models by employing a faster assistant model. The assistant model generates a sequence of candidate tokens, all of which are verified by the main model in a single forward pass. By generating with the faster assistant model and only performing validation forward passes with the main model, the decoding process is sped-up significantly.
The $i$-th candidate token from the assistant model $\hat{y}_{i}$ is only kept if all previous candidate tokens $\hat{\boldsymbol{y}}_{<i}$ match the validation tokens predicted by the main model. Consequently, speculative decoding ensures that the generated output exactly matches the sequence of tokens that would be generated by the main model, making it a natural replacement for existing inference pipelines that use the main model.

DistilSpec \citep{zhou23_distilspec} proposes using a knowledge-distilled student model as the assistant to better align the distribution of the assistant model with the main one. We apply the same principal here, using Distil-Whisper as the assistant to Whisper.

\section{Experimental Setup}
\subsection{Data}

Inspired by SpeechStew \citep{chan21_speechstew}, we assemble a large corpus of ASR training data for large-scale KD through a combination of nine publicly available speech recognition datasets. An overview of the datasets is presented in Table~\ref{tab:datasets-summary}, with additional details in Appendix~\ref{sec: appendix-training-data}. The combined dataset contains 21,170 hours of speech data, encompassing over 18,260 speakers and 10 distinct domains. We load and pre-process all datasets in the Hugging Face Datasets library \citep{lhoest21_datasets}, streaming the data from the Hugging Face Hub\footnote{Training datasets: \url{https://huggingface.co/collections/distil-whisper/training-datasets-6538d05c69721489d1db1e49}}.

\begin{table}[t]
\caption{Summary of the open-source datasets used for training. For some datasets, the number of speakers cannot be reliably retrieved. We denote these entries as ``unknown".}
\begin{center}
\label{tab:datasets-summary}
\begin{tabular}{@{}lrrll@{}}
\toprule
\textbf{Dataset}         & \multicolumn{1}{l}{\textbf{Size / h}} & \multicolumn{1}{l}{\textbf{Speakers}} & \textbf{Domain}                      & \textbf{Licence}         \\ \midrule
People's Speech & 12,000                        & unknown                      & Government, interviews     & CC-BY-SA-4.0    \\
GigaSpeech      & 2,500                         & unknown                      & Audiobook, podcast, YouTube & apache-2.0      \\
Common Voice 13           & 2,400                         & unknown                      & Narrated Wikipedia               & CC0-1.0         \\
Fisher          & 1,960                         & 11,900                         & Telephone conversations     & LDC             \\
LibriSpeech     & 960                          & 2,480                        & Audiobooks                  & CC-BY-4.0       \\
VoxPopuli       & 540                          & 1,310                         & European Parliament         & CC0             \\
TED-LIUM        & 450                           & 2,030                         & TED talks                   & CC-BY-NC-ND 3.0 \\
SwitchBoard            & 260                           & 540                          & Telephone conversations     & LDC             \\
AMI             & 100                           & unknown                      & Meetings                    & CC-BY-4.0       \\ \midrule
Total           & 21,170                        & 18,260+                      &                             &                 \\ \bottomrule
\end{tabular}
\end{center}
\end{table}

We generate pseudo-labels for our training data with the Whisper large-v2 checkpoint, using the Flax Whisper implementation in the Hugging Face Transformers library \citep{heek20_flax, wolf20_transformers}. We found there to be little difference in the downstream performance of the distilled model after pseudo-labelling using either greedy or beam-search, and so we opted to pseudo-label the training data with greedy decoding for its faster inference speed.

\subsection{Training}

\begin{figure}
    \begin{center}
    \includegraphics[width=0.8\textwidth]{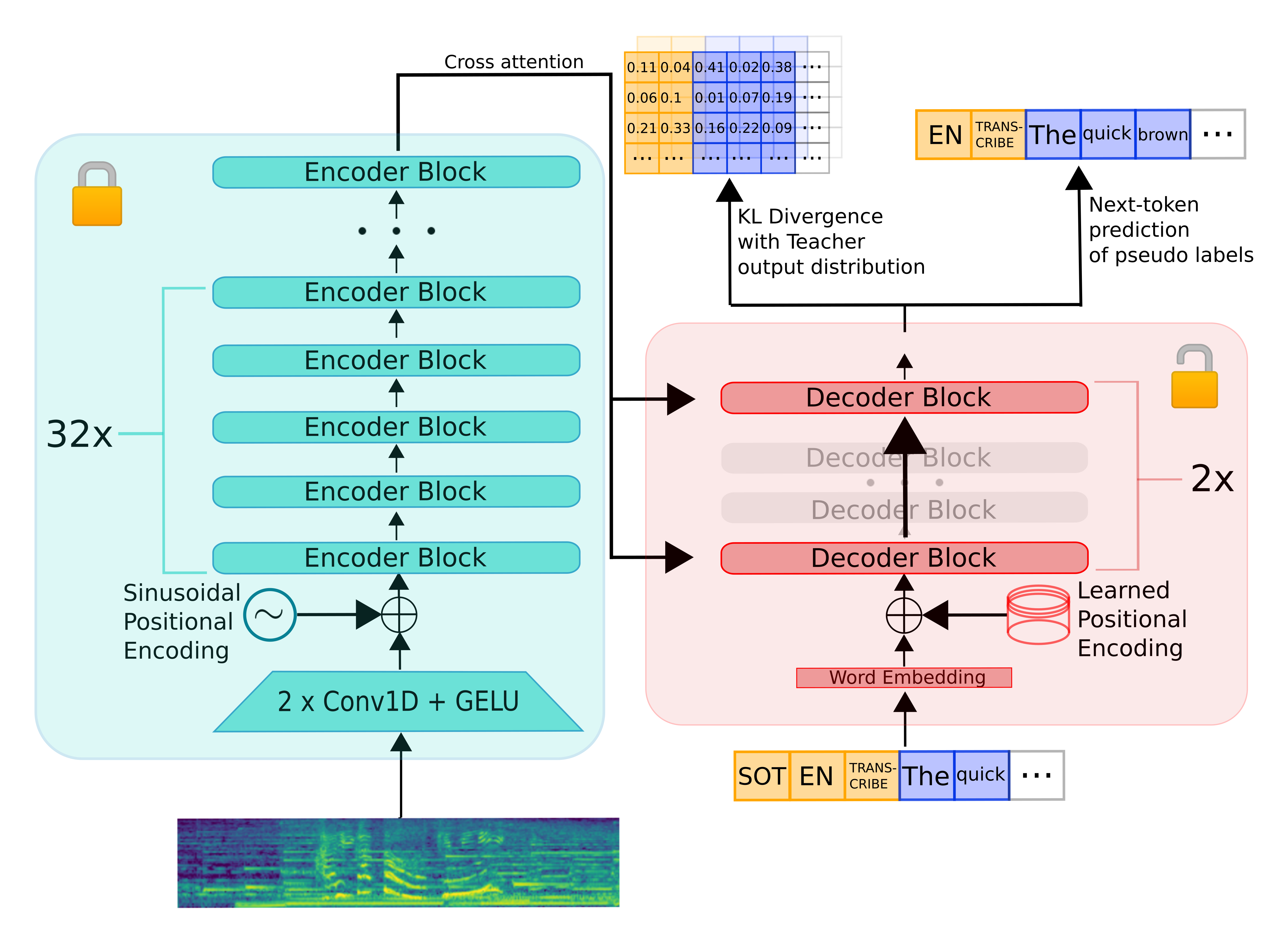}
    \caption{\textbf{Architecture of the Distil-Whisper model.} The encoder (shown in green) is entirely copied from the teacher to the student and frozen during training. The student's decoder consists of only two decoder layers, which are initialised from the first and last decoder layer of the teacher (shown in red). All other decoder layers of the teacher are discarded. The model is trained on a weighted sum of the KL divergence and PL loss terms.}
    \label{fig: arch}
    \end{center}
\end{figure}

We initialise the student models by copying the entire encoder from the teacher and freeze it during training. We distill 2-layer decoder checkpoints from the medium.en and large-v2 models by copying the first and last decoder layers, which we refer to as distil-medium.en and distil-large-v2 respectively. The dimensionality details of the distilled models are shown in Table~\ref{tab: whisper distil checkpoints}, with the architecture and training objective summarised in Figure~\ref{fig: arch}.

We train with a batch size of 256 for a total of 80,000 optimisation steps, which amounts to eight epochs of training. Since we only train for eight epochs, the risk of over-fitting is low, and so we do not use any data augmentation or regularisation techniques. Instead, we rely on the diversity of our dataset to ensure model generalisation and robustness, the same premise used in training the original Whisper model \citep{radford22_whisper}. Refer to Appendix~\ref{sec:appendix-training} for full details of our training set-up.

\begin{table}
\caption{Dimensionality details of Distil-Whisper checkpoints.}
\begin{center}
\label{tab: whisper distil checkpoints}
\begin{tabular}{@{}lrrrrr@{}}
\toprule
\textbf{Model}     & \multicolumn{1}{l}{\textbf{Enc. Layers}} & \multicolumn{1}{l}{\textbf{Dec. Layers}} & \multicolumn{1}{l}{\textbf{Width}} & \multicolumn{1}{l}{\textbf{Heads}} & \multicolumn{1}{l}{\textbf{Params. / M}} \\ \midrule
\href{https://huggingface.co/distil-whisper/distil-medium.en}{distil-medium.en}     & 24 & 2                          & 1024                       & 16                         & 394                                 \\
\href{https://huggingface.co/distil-whisper/distil-large-v2}{distil-large-v2}     & 32 & 2                          & 1280                       & 20                         & 756                                                            \\ \bottomrule
\end{tabular}
\end{center}
\end{table}

\subsection{Short-Form Evaluation}

The objective of Distil-Whisper is to compress the original Whisper model into a smaller, faster variant of the model that retains its robustness to different acoustic conditions (speakers, speaking styles and domains). To investigate this capability, we employ a broad collection of speech recognition datasets to examine whether Distil-Whisper can effectively generalise across datasets and domains.

We evaluate the Distil-Whisper model on a total of 15 short-form datasets. The first 11 of these datasets are the corresponding test splits for the training data used to distil the model. These test splits are \textit{in distribution (ID)} with the training data. The remaining four datasets are used as test sets only, where Distil-Whisper is assessed in a zero-shot setting, that is without the use of the corresponding training data, thereby measuring the model's ability to generalise to \textit{out of distribution (OOD)} datasets. An overview of the OOD evaluation datasets is presented in Table~\ref{tab: eval datasets}. Full details of the evaluation datasets are provided in Appendix~\ref{sec: appendix-short-form-data}.

We examine both overall robustness, that is the average performance over all datasets, and effective robustness \citep{taori20_robustness}, which measures the difference in expected performance between a reference dataset that is ID, and one or more datasets that are OOD. A model with high effective robustness does better on OOD datasets as a function of performance on the reference dataset. A model with ideal effective robustness performs equally well on all datasets. In our experiments, we use GigaSpeech \citep{chen21_gigaspeech} as the reference dataset, owing to the fact it contains web-scraped data from audiobooks, podcasts and YouTube videos, and is such ID with both the pre-trained Whisper training data and the distilled Whisper train set.

\begin{table}
\caption{Summary of the OOD datasets used for short and long-form evaluation.}
\label{tab: eval datasets}
\begin{center}
\begin{tabular}{lrrll} 
\toprule
\textbf{Dataset} & \multicolumn{1}{l}{\textbf{Size / h}} & \multicolumn{1}{l}{\textbf{Speakers}} & \textbf{Domain}    & \textbf{Licence}  \\ 
\midrule
\multicolumn{5}{c}{\textit{Short-Form}}                                                                                                   \\ 
\midrule
CHiME-4          & 7                                     & 87                                    & News broadcast     & LDC               \\
Earnings-22      & 115                                   & unknown                               & Financial meetings & CC-BY-SA-4.0      \\
FLEURS           & 2                                     & 3                                     & Narrated Wikipedia & CC-BY-4.0         \\
SPGISpeech       & 100                                   & unknown                               & Financial meetings & User Agreement    \\ 
\midrule
\multicolumn{5}{c}{\textit{Long-Form}}                                                                                                    \\ 
\midrule
Earnings-21      & 39                                    & unknown                               & Financial meetings & CC-BY-SA-4.0      \\
Earnings-22      & 115                                   & unknown                               & Financial meetings & CC-BY-SA-4.0      \\
Meanwhile        & 1                                     & 1                                     & TV show            & User Agreement              \\
Rev 16           & 16                                    & unknown                               & Podcasat           & CC-BY-4.0              \\
\bottomrule
\end{tabular}
\end{center}
\end{table}

We evaluate the noise robustness of the Distil-Whisper models, the original Whisper models, and eight other LibriSpeech-trained models by measuring the WER on the LibriSpeech test-clean dataset with increasing amounts of noise applied to the input audio. The LibriSpeech dataset is an ideal choice of dataset since it has a high signal-to-noise ratio (SNR), and thus enablers evaluation over a large range of SNRs as the amount of noise is increased. We add either white noise or pub noise from the Audio Degradation Toolbox \citep{matthias13_toolbox}. The pub noise simulates a naturally noisy environment, with ambient sounds and indistinguishable conversations characteristic of a busy restaurant or pub. The level of additive noise is determined based on the signal power of individual instances, and corresponds to a specified SNR.

\subsection{Long-Form Evaluation}

We evaluate the long-form transcription performance of the Distil-Whisper model on four OOD datasets comprising different lengths and acoustic conditions, in order to cover the broadest possible distribution of data. An overview of the long-form datasets is presented in Table~\ref{tab: eval datasets}. Full details about the long-form datasets are provided in Appendix~\ref{sec: appendix-long-form-data}.

The Whisper model demonstrates a susceptibility to hallucinate, characterised by either the repetitive generation of identical sequences, or predicting passages of text not spoken in the audio input \citep{bain23_whisperx, zhang23_usm}. These hallucinations errors are most prevalent in long-form audio transcription, particularly when the audio contains large amounts of silence between spoken utterances. 
To quantify the amount of repetition and hallucination in the predicted transcriptions, we measure the number of repeated 5-gram word duplicates (5-Dup.) and the insertion error rate (IER) over the four OOD long-form datasets. We also report the substitution error rate (SER) and deletion error rate (DER) to quantify the frequency of substitutions and deletions in the transcriptions.

\section{Results}
\subsection{Short-Form Evaluation}

Table~\ref{tab: overall results} reports the average WER scores over the four OOD short-form test sets for the Whisper and Distil-Whisper checkpoints. For a detailed breakdown of results on a per-dataset basis, refer to Appendix~\ref{sec:appendix-eval}. Of the two distilled models, the distil-large-v2 model achieves the lowest overall average WER of 10.1\%. It is one percentage point higher than the large-v2 baseline, with 5.8 times faster inference speed and fewer than half the parameters. The inference speed is comparable to the tiny.en Whisper checkpoint, but with an 8.8\% WER advantage. The distil-medium.en model is on average 2.0\% WER higher than the large-v2 model, with 6.8x faster inference and 75\% model compression. These findings highlight that Distil-Whisper retains the overall robustness of the Whisper model, with comparable WER performance averaged over multiple OOD datasets, but with significantly faster inference speed and reduced parameter count.

\begin{table}
\caption{\textbf{Distil-Whisper retains the WER performance of the Whisper model but with faster inference speed.} Average WER results over the four OOD short-form test sets and the four OOD long-form test sets. Relative latency is the inference time relative to the large-v2 checkpoint. 
For short-form evaluation, the batch size is set to 1. For long-form evaluation, the chunked long-form transcription algorithm is used with a batch size of 16.}
\label{tab: overall results}
\begin{center}
\begin{tabular}{l|r|rr|rr} 
\toprule
\multirow{2}{*}{\textbf{Model}} & \multicolumn{1}{l|}{\multirow{2}{*}{\textbf{Params / M}}} & \multicolumn{2}{c|}{\textbf{Short Form}}                         & \multicolumn{2}{c}{\textbf{Long Form}}                           \\
                                & \multicolumn{1}{l|}{}                                     & \multicolumn{1}{l}{Rel. Latency} & \multicolumn{1}{l|}{Avg. WER} & \multicolumn{1}{l}{Rel. Latency} & \multicolumn{1}{l}{Avg. WER}  \\ 
\midrule
\href{https://huggingface.co/openai/whisper-tiny.en}{tiny.en}                            & \textbf{39}                                               & 6.1                              & 18.9                          & 5.4                              & 18.9                          \\
\href{https://huggingface.co/openai/whisper-base.en}{base.en}                            & 74                                                        & 4.9                              & 14.3                          & 4.3                              & 15.7                          \\
\href{https://huggingface.co/openai/whisper-small.en}{small.en}                           & 244                                                       & 2.6                              & 10.8                          & 2.2                              & 14.7                          \\
\href{https://huggingface.co/openai/whisper-medium.en}{medium.en}                          & 769                                                       & 1.4                              & 9.5                           & 1.3                              & 12.3                          \\
\href{https://huggingface.co/openai/whisper-large-v2}{large-v2}                        & 1550                                                      & 1.0                              & \textbf{9.1}                  & 1.0                              & 11.7                          \\ 
\midrule
\href{https://huggingface.co/distil-whisper/medium.en}{distil-medium.en}                     & 394                                                       & \textbf{6.8}                     & 11.1                          & \textbf{8.5}                     &     12.4                         \\
\href{https://huggingface.co/distil-whisper/large-v2}{distil-large-v2}                   & 756                                                       & 5.8                              & 10.1                          & 5.8                              & \textbf{11.6}                 \\
\bottomrule
\end{tabular}
\end{center}
\end{table}

Table~\ref{tab: rer eval results} compares the effective robustness of large-v2 to distil-large-v2. 
The models have very close performance on the reference distribution, performing to within 2\% relative WER. The distilled model improves upon the pre-trained baseline for the SPGISpeech dataset by 12.8\% relative, but performs worse on the three other OOD datasets. Compared to the teacher model, the distilled model achieves an overall WER increase of 0.8\% absolute (or 10.7\% relative). The narrow performance gap indicates that Distil-Whisper has comparable effective robustness to the original Whisper model.

\begin{table}
\caption{\textbf{Effective robustness across various datasets.} WER results for one ID reference dataset and four OOD datasets. The relative error rate (RER) is shown on the right, giving the per-dataset effective robustness scores. The macro-average results are shown in the bottom row.}
\label{tab: rer eval results}
\begin{center}
\begin{tabular}{l|rr|r}
\toprule
\textbf{Dataset}            & \multicolumn{1}{l}{\textbf{\href{https://huggingface.co/openai/whisper-large-v2}{large-v2}}} & \multicolumn{1}{l|}{\textbf{\href{https://huggingface.co/distil-whisper/large-v2}{distil-large-v2}}} & \multicolumn{1}{l}{\textbf{RER}}  \\ 
\midrule
GigaSpeech  & 10.7                         & 10.5                            & -2.0                     \\
\midrule
CHIME-4     & 11.8                         & 14.0                            & 18.4                     \\
Earnings-22 & 16.6                         & 16.9                            & 1.6                      \\
FLEURS      & 4.2                          & 6.3                             & 48.2                     \\
SPGISpeech  & 3.8                          & 3.3                             & -12.8                    \\ 
\midrule
Average     & 9.4                          & 10.2                            & 10.7                      \\
\bottomrule
\end{tabular}
\end{center}
\end{table}

\subsection{Long-Form Evaluation} \label{sec: long-form eval}

We compare the long-form transcription performance of Distil-Whisper to the pre-trained Whisper models on the four OOD long-form test sets. Table~\ref{tab: overall results} reports the relative latency for a batch size of 16, as well as the macro-average WER. The per-dataset WER scores are provided in Appendix~\ref{sec:appendix-eval}. The results show that distil-large-v2 outperforms or equals large-v2 on four of the five test sets, with an average WER that is 0.1\% lower with 5.8 times faster batched inference speed. It is only on the Meanwhile dataset that the Distil-Whisper model performs worse, which contains recordings from a single speaker with a high-frequency of uncommon words.

Table~\ref{tab: long form algorithm} compares the performance of the Whisper long-form sequential algorithm to the Distil-Whisper chunked one. The large-v2 model with the chunked algorithm yields a 9.9 times speed-up compared to the sequential one, with a 1.3\% increase to average OOD WER. This demonstrates the inference speed gain that is achieved through batching. Using the distilled model in combination with the chunked algorithm provides further improvements: the distil-large-v2 model is 57.5 times faster than the baseline large-v2 implementation, while performing to within 1.2\% WER. 

\begin{table}
\caption{\textbf{Comparison of long-form transcription algorithms.} Average WER results over the four OOD long-form test sets for the sequential and chunked long-form algorithms. Relative latency is the inference time relative to the large-v2 model with sequential long-form decoding. The chunked transcription results are reported using a batch size of 16.}
\label{tab: long form algorithm}
\begin{center}
\begin{tabular}{llrr} 
\toprule
\textbf{Model}    & \textbf{Algorithm} & \multicolumn{1}{l}{\textbf{Rel. Latency}} & \multicolumn{1}{l}{\textbf{Avg. OOD WER}}  \\ 
\midrule
\href{https://huggingface.co/openai/whisper-large-v2}{large-v2}         & Sequential        & 1.0                                       & \textbf{10.4}                     \\
\href{https://huggingface.co/openai/whisper-large-v2}{large-v2}          & Chunked      & 9.9                                       & 11.7                              \\
\href{https://huggingface.co/distil-whisper/large-v2}{distil-large-v2} & Chunked      & \textbf{57.5}                             & 11.6                              \\
\bottomrule
\end{tabular}
\end{center}
\end{table}

\subsection{Robustness to Additive Noise}

\begin{figure}
    \begin{center}
    \includegraphics[width=\textwidth]{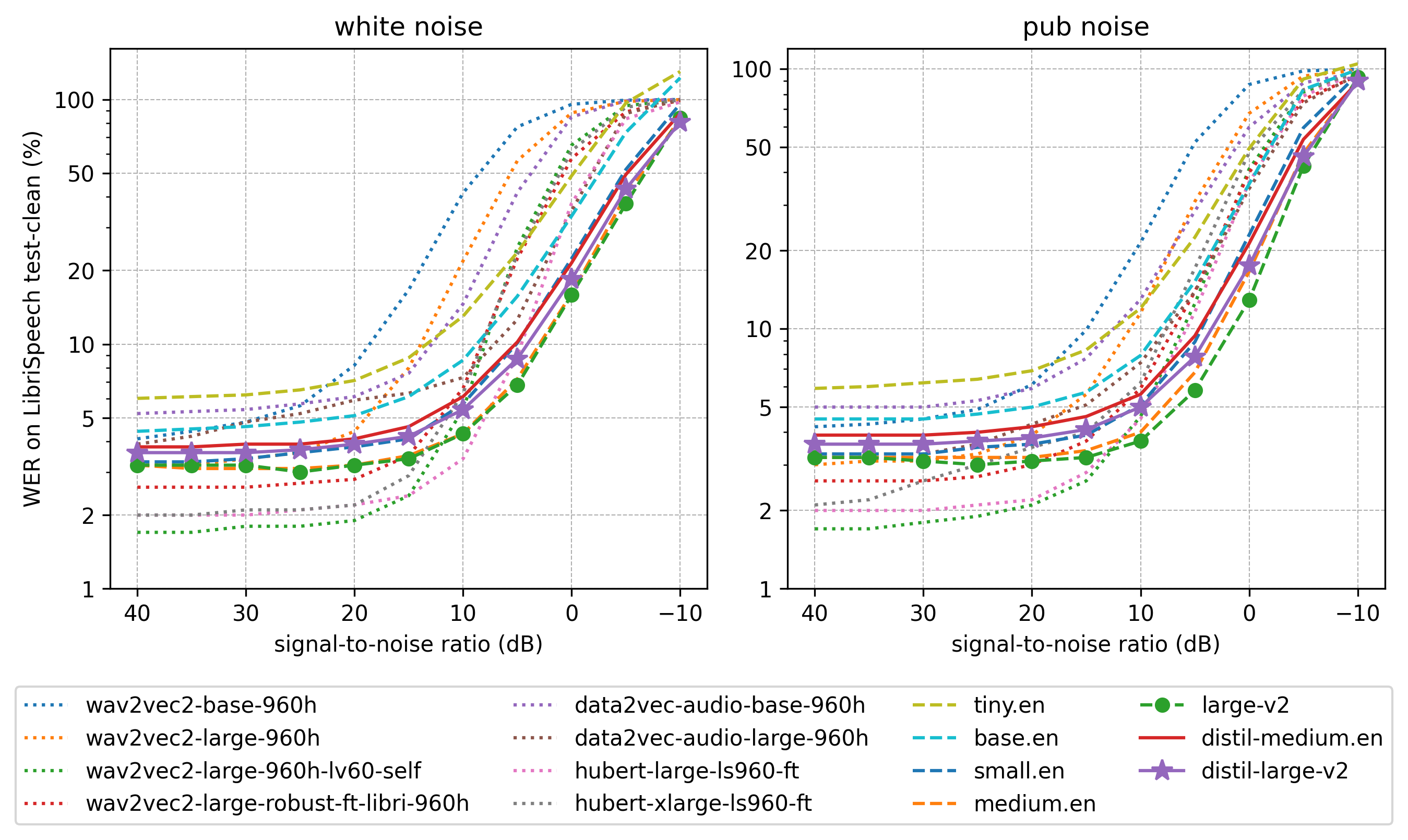}
    \caption{\textbf{Effect of noise on WER performance.} WER on LibriSpeech test-clean as a function of SNR under additive white noise (left) and pub noise (right).}
    \label{fig: snr}
    \end{center}
\end{figure}

Figure~\ref{fig: snr} shows how WER performance degrades as the intensity of additive noise increases on the LibriSpeech test-clean dataset. Of the 14 models we compare to, eight are pre-trained and/or fine-tuned on LibriSpeech. There are many models that outperform the Distil-Whisper models under low noise (40 dB SNR), including the Whisper checkpoints and LibriSpeech trained models. However, as the noise becomes more intensive, the WERs of the Distil-Whisper checkpoints degrade less severely than the LibriSpeech trained models and approach those of Whisper, especially when the additive pub noise decreases below 10 dB. Since we copy the full encoder and freeze it during training, the student and teacher models share the same encoder. Thus, they show similar robustness to noise, particularly under more natural distribution shifts like pub noise.

\subsection{Robustness to Hallucinations}

Table~\ref{tab: hallucinations} reports the number of repeated 5-gram word duplicates (5-Dup.) and insertion error rate (IER) metrics averaged over the four long-form test sets. In addition to the Whisper and Distil-Whisper models, we report the results for the official Wav2Vec 2.0 large model fine-tuned on 960 hours of LibriSpeech data. This checkpoint provides a comparison between the Whisper Seq2Seq architecture and a CTC based one \citep{graves06_ctc}. A CTC model should be less prone to hallucination errors given it is an encoder-only architecture with a linear head over the vocabulary.

The distil-large-v2 model has 1.3 times fewer repetition errors than Whisper large-v2. It also obtains the lowest average IER, improving on Whisper by 1.2\% absolute. This indicates that the amount of hallucination is improved in Distil-Whisper compared to the original Whisper model. The average deletion error rate (DER) is comparable for both large-v2 and distil-large-v2, performing to within 0.3\% DER. However, the substitution error rate (SER) is 1.4\% higher for distil-large-v2, indicating that the distilled models are subject to more substitution errors. Overall, the reduction in IER outweighs the increase to SER, and Distil-Whisper returns the lowest WER of all the models. While the wav2vec 2.0 model underperforms in its average WER score, we find that it is far less prone to repetition errors compared to both Whisper and Distil-Whisper. Further work is needed to reduce repetition errors in Seq2Seq ASR models.

\begin{table}
\caption{\textbf{Detailed long-form error rates.} Average number of repeated 5-gram word duplicates (5-Dup.) and insertion error rate (IER) over the four long-form test sets. Shown also are the average substitution error rate (SER), deletion error rate (DER) and word error rate (WER) metrics.}
\label{tab: hallucinations}
\begin{center}
\begin{tabular}{lrrrrr} 
\toprule
\textbf{Model} & \multicolumn{1}{l}{\textbf{5-Dup.}} & \multicolumn{1}{l}{\textbf{IER}} & \multicolumn{1}{l}{\textbf{SER}} & \multicolumn{1}{l}{\textbf{DER}} & \multicolumn{1}{l}{\textbf{WER}}  \\ 
\midrule
\href{https://huggingface.co/facebook/wav2vec2-large-960h}{wav2vec2-large-960h}               & \textbf{7971}                       & 4.8                              & 18.9                             & 4.6                              & 28.3                              \\ 
\midrule
\href{https://huggingface.co/openai/whisper-tiny.en}{tiny.en}               & 23313                               & 5.1                              & 8.9                              & 4.8                              & 18.9                              \\
\href{https://huggingface.co/openai/whisper-base.en}{base.en}                & 22719                               & 4.3                              & 6.6                              & 4.8                              & 15.7                              \\
\href{https://huggingface.co/openai/whisper-small.en}{small.en}                & 26377                               & 3.3                              & 5.0                              & 6.5                              & 14.7                              \\
\href{https://huggingface.co/openai/whisper-medium.en}{medium.en}               & 23549                               & 3.5                              & 4.2                              & 4.6                              & 12.3                              \\
\href{https://huggingface.co/openai/whisper-large-v2}{large-v2}                & 23792                               & 3.3                              & \textbf{3.9}                     & 4.5                              & 11.7                              \\ 
\midrule
\href{https://huggingface.co/distil-whisper/medium.en}{distil-medium.en}               & 18918                               & 2.5                              & 5.6                              & 4.4                              & 12.4                              \\
\href{https://huggingface.co/distil-whisper/large-v2}{distil-large-v2}               & 18503                               & \textbf{2.1}                     & 5.3                              & \textbf{4.2}                     & \textbf{11.6}                     \\
\bottomrule
\end{tabular}
\end{center}
\end{table}

\subsection{Speculative Decoding}

Table~\ref{tab: spec dec} reports the relative latency of the medium.en and large-v2 models with speculative decoding. We compare the latency using either the smallest Whisper checkpoints or the Distil-Whisper models as the assistant. Since the outputs of the original Whisper models are obtained exactly, we report the relative latency only. The distilled student models are initialised by copying and freezing the entire encoder from the teacher, meaning they use exactly the same encoder as main Whisper models. Therefore, when running SD, the encoder can be shared between the main and assistant models, and only the distilled decoder layers have to be loaded in addition to the main model. This results in just an 8\% increase to parameter count when using distil-large-v2 as the assistant to large-v2. 

Speculative decoding with the distil-large-v2 assistant yields a 2.0 times improvement to inference speed over large-v2 alone. This is comparable to using the tiny model as the assistant. For the medium.en model, using distil-medium.en as an assistant provides a 2.4 times speed-up. This is greater than using the tiny.en checkpoint as an assistant, which is only 2.0 times faster. Overall, speculative decoding provides significant speed-ups to latency while mathematically ensuring the same outputs, making it a natural replacement for existing Whisper pipelines.

\begin{table}
\caption{\textbf{Impact of speculative decoding.} Relative latency of medium.en and large-v2 using Whisper and Distil-Whisper assistant models. The relative latency is computed relative to the large-v2 model without speculative decoding for a batch size of 1.}
\label{tab: spec dec}
\begin{center}
\begin{tabular}{lrr}
\toprule
\textbf{Model}    & \multicolumn{1}{l}{\textbf{Params / M}} & \multicolumn{1}{c}{\textbf{Rel. Latency}}  \\ 
\midrule
\href{https://huggingface.co/openai/whisper-medium.en}{medium.en}                          & 769                                     & 1.4                                        \\
\text{  with} \href{https://huggingface.co/openai/whisper-tiny.en}{tiny.en}           & 808                                     & 2.7                                        \\
\text{  with} \href{https://huggingface.co/distil-whisper/medium.en}{distil-medium.en} & 856                                     & 3.3                                        \\ 
\midrule
\href{https://huggingface.co/openai/whisper-large-v2}{large-v2}                            & 1550                                    & 1.0                                        \\
\text{  with} \href{https://huggingface.co/openai/whisper-tiny}{tiny}                  & 1589                                    & 2.1                                        \\
\text{  with} \href{https://huggingface.co/distil-whisper/large-v2}{distil-large-v2}   & 1669                                    & 2.0                                        \\
\bottomrule
\end{tabular}
\end{center}
\end{table}

\section{Analysis}
\subsection{WER Threshold} \label{sec: wer threshold analysis}

During training, we filter pseudo-labelled data where the normalised WER between the Whisper-generated pseudo-labels and the ground truth labels exceeds a given threshold $\lambda$. To investigate the effect of this threshold on the performance of the distilled model, we train a series of distil-large-v2 checkpoints for 10,000 training steps (or two epochs) on a range of threshold values. Table~\ref{tab: wer threshold} shows the average WER performance of the trained models. Setting the threshold too high allows mis-transcribed or hallucinated transcriptions to enter the training set. Setting the WER threshold low retains only the most accurate Whisper-generated pseudo-labels, but also only the easiest samples (i.e. those with very low WER) and discards a larger proportion of the training data. We found that a WER threshold of 10\% provides a good trade-off between these opposing factors. Had we trained for longer, the effect of using a higher quantity of training data might have become more pronounced, thus favouring higher thresholds. Using a WER threshold to filter pseudo-labelled data may compensate for the decreased transcription accuracy of the Whisper-generated labels predicted with greedy decoding as opposed to beam-search. We find it is an effective strategy for improving the performance of Seq2Seq ASR systems trained on pseudo-labelled data.

\begin{table}
\caption{\textbf{The WER threshold is an effective filter for PL data.} Average WER of the distil-large-v2 checkpoint on the 11 ID and three OOD validation sets as the WER threshold $\lambda$ is reduced.}
\label{tab: wer threshold}
\begin{center}
\begin{tabular}{rrrrr} 
\toprule
\multicolumn{1}{l}{$\lambda$} & \multicolumn{1}{l}{\textbf{Data Filtered / \%}} & \multicolumn{1}{l}{\textbf{Avg. ID WER}} & \multicolumn{1}{l}{\textbf{Avg. OOD WER}} & \multicolumn{1}{l}{\textbf{Avg. WER}}  \\ 
\midrule
100                                 & 0.0                                             & 14.8                                     & 9.1                                       & 13.4                                   \\
80                                  & 6.2                                             & 13.5                                     & 7.5                                       & 12.1                                   \\
40                                  & 11.9                                            & 13.3                                     & 7.4                                       & 12.0                                   \\
20                                  & 24.2                                            & 13.1                                     & 7.3                                       & 11.7                                   \\
15                                  & 32.0                                            & 13.0                                     & 7.4                                       & 11.7                                   \\
10                                  & 45.4                                            & 12.6                                     & 7.4                                       & 11.4                                   \\
5                                   & 60.3                                            & 12.6                                     & 7.3                                       & 11.4                                   \\
\bottomrule
\end{tabular}
\end{center}
\end{table}

\subsection{Dataset Scaling}

To study the amount of data is required to distil Whisper, we trained a series of distil-large-v2 models on subsampled versions of our dataset. 
Table~\ref{tab: dataset scaling} shows the average WER performance of the distil-large-v2 model for each of the dataset proportions. All increases in dataset size result in improved performance on the ID validation sets. On the OOD validation sets, performance improves rapidly from 435 to 3,483 hours, and then slows down significantly between 3,483 hours and 13,933 hours. Using the full dataset of 21,770 hours -- a further 1.6 times increase in size -- results in no further improvement to OOD WER. This suggests that there are diminishing gains increasing the amount of pseudo-labelled training data above 13,933 hours, or 2\% of the Whisper pre-training data.


\begin{table}
\caption{\textbf{Performance improves with increasing dataset size.} Average WER of the distil-large-v2 checkpoint on the 11 ID and three OOD validation sets as the amount of training data is increased.}
\label{tab: dataset scaling}
\begin{center}
\begin{tabular}{rrrrr} 
\toprule
\multicolumn{1}{l}{\textbf{Size / h}} & \multicolumn{1}{l}{\textbf{Proportion / \%}} & \multicolumn{1}{l}{\textbf{Avg. ID WER}} & \multicolumn{1}{l}{\textbf{Avg. OOD WER}} & \multicolumn{1}{l}{\textbf{Avg. WER}}  \\ 
\midrule
435                                   & 2                                            & 17.1                                     & 13.8                                      & 16.4                                   \\
871                                   & 4                                            & 15.1                                     & 10.5                                      & 14.0                                   \\
1,742                                 & 8                                            & 14.0                                     & 9.2                                       & 12.9                                   \\
3,483                                 & 16                                           & 13.3                                     & 7.8                                       & 12.0                                   \\
6,966                                 & 32                                           &         13.0	                            &     7.7	                                  &       11.8                              \\
13,933                                & 64                                           & 12.8                                     & 7.4                                       & 11.6                                   \\
21,770                                & 100                                          & 12.6                                     & 7.4                                       & 11.4                                   \\
\bottomrule
\end{tabular}
\end{center}
\end{table}

\subsection{Model Size}


Table~\ref{tab: model size} shows the latency and WER performance for 16, 8, 4 and 2 decoder-layer models. The 16-layer decoder model largely retains the WER performance of the 32-layer teacher model, performing to within 0.1\% OOD WER with a 1.9 times improvement in latency. As the number of decoder layers is reduced further, the average OOD WER increases more significantly. The maximum increase is 2.1\% for the 2-layer decoder model, however this configuration is 5.8 times faster than the teacher model. These results highlight the trade-off between latency and performance as the number of decoder layers is reduced.


\begin{table}
\caption{\textbf{Trade-off between latency and WER performance with decreasing model size.} Average WER over the 11 ID and three OOD validation sets as the number of encoder and decoder layers in the large-v2 checkpoint are reduced. The first row corresponds to the teacher checkpoint large-v2. The following rows correspond to the distilled models, which are trained for 10,000 optimisation steps (or two epochs).}
\label{tab: model size}
\begin{center}
\begin{tabular}{rrrrrrr} 
\toprule
\multicolumn{1}{l}{\textbf{Enc}} & \multicolumn{1}{l}{\textbf{Dec}} & \multicolumn{1}{l}{\textbf{Params / M}} & \multicolumn{1}{l}{\textbf{Rel. Latency}} & \multicolumn{1}{l}{\textbf{ID WER}} & \multicolumn{1}{l}{\textbf{OOD WER}} & \multicolumn{1}{l}{\textbf{Avg. WER}}  \\ 
\midrule
32                               & 32                               & 1543                                    & 1.0                                       & 12.8                                & 5.3                                  & 11.1                                   \\ 
\midrule
32                               & 16                               & 1124                                    & 1.9                                       & 11.3                                & 5.4                                  & 9.9                                    \\
32                               & 8                                & 914                                     & 3.0                                       & 11.6                                & 6.0                                  & 10.3                                   \\
32                               & 4                                & 809                                     & 4.3                                       & 12.0                                & 6.5                                  & 10.8                                   \\
32                               & 2                                & 756                                     & 5.8                                       & 12.6                                & 7.4                                  & 11.4                                   \\ 
\midrule
16                               & 2                                & 441                                     & 8.6                                       &            16.0	                        &               10.5	                       &              14.7                          \\
\bottomrule
\end{tabular}
\end{center}
\end{table}

In distilling only the decoder, the parameter reduction is limited to 51\%. To further reduce the parameter count, we can jointly distil the encoder and decoder. Table~\ref{tab: model size} compares the performance of a full 32-layer encoder student model to a reduced 16-layer one, both with 2-layer decoders. When the encoder is reduced to 16-layers, the parameter count decreases by an additional 19\%. However, the OOD WER performance degrades by 3.1\% absolute. This suggests that having a deep encoder is paramount for maintaining strong WER performance of distilled Seq2Seq ASR models.

\section{Conclusion}

We introduce Distil-Whisper, a distilled version of Whisper that is 49\% smaller, 5.8 times faster, and within 1\% WER performance on OOD short-form audio. On OOD long-form audio, Distil-Whisper outperforms Whisper, due to fewer hallucinations and repetitions. We show that large-scale pseudo-labelling is an effective strategy for distilling ASR models, in particular when combined our WER threshold filter. We further demonstrate that Distil-Whisper can be used in combination with Whisper using speculative decoding to obtain the same outputs as the original model with 2 times faster inference.

\section{Acknowlegements}

We thank Nicolas Patry and Arthur Zucker for their implementation of the chunked long-form transcription algorithm in Transformers and João Gante for the implementation of speculative decoding. We gratefully acknowledge the support of Google’s TPU Research Cloud (TRC) program for providing Cloud TPU resources for this research.

\bibliography{bibliography}
\bibliographystyle{iclr2023_conference}

\appendix
\section{Additional Dataset Details} \label{sec:appendix-data-preparation}
\subsection{Training Data} \label{sec: appendix-training-data}

Quantitative and qualitative information about the training datasets is displayed in Tables~\ref{tab: training-quant} and~\ref{tab: training-qual} respectively. A detailed description of the training datasets is presented below.

\begin{table}[b]
\caption{\textbf{Quantitative statistics of the training datasets.} The mean audio length is quoted in seconds, and the mean transcription length in number of words.}
\label{tab: training-quant}
\begin{center}
\begin{tabular}{lrrrr} 
\toprule
\textbf{Dataset} & \multicolumn{1}{l}{\textbf{Size / h}} & \multicolumn{1}{l}{\textbf{Speakers}} & \multicolumn{1}{l}{\textbf{Mean Audio (s)}} & \multicolumn{1}{l}{\textbf{Mean Text (words)}}  \\ 
\midrule
People's Speech  & 12,000                                & unknown                               & 13.8                                        & 38.5                                     \\
Common Voice 13  & 3,000                                 & unknown                               & 5.6                                         & 31.9                                     \\
GigaSpeech       & 2,500                                 & unknown                               & 4.0                                         & 12.9                                     \\
Fisher           & 1,960                                 & 11,900                                & 3.3                                         & 10.1                                      \\
LibriSpeech      & 960                                   & 2,480                                 & 12.1                                        & 32.9                                     \\
VoxPopuli        & 540                                   & 1,310                                 & 10.3                                        & 26.1                                     \\
TED-LIUM         & 450                                   & 2,030                                 & 6.1                                         & 18.3                                     \\
SwitchBoard      & 260                                   & 540                                   & 4.8                                         & 8.3                                      \\
AMI              & 100                                   & unknown                               & 2.6                                         & 7.3                                      \\ 
\midrule
Total            & 21,770                                & 18,260+                               & 7.1                                         & 19.8                                     \\
\bottomrule
\end{tabular}
\end{center}
\end{table}

\begin{table}[t]
\caption{\textbf{Qualitative statistics of the training datasets.} The speaking styles are narrated (N), oratory (O) or spontaneous (S), or a combination of them.}
\label{tab: training-qual}
\begin{center}
\begin{tabular}{lllll} 
\toprule
\textbf{Dataset} & \textbf{Domain}             & \textbf{Rec. Cond.} & \textbf{Style} & \textbf{Licence}  \\ \midrule
People's Speech  & Internet Archive            & Close-talk mic.     & N, O, S        & CC-BY-SA-4.0      \\
Common Voice 13  & Narrated Wikipedia          & Close-talk mic.     & N              & CC0-1.0           \\
GigaSpeech       & Audiobook, podcast, YouTube & Close-talk mic.     & N, S           & apache-2.0        \\
Fisher           & Telephone conversations     & Telephone           & S              & LDC               \\
LibriSpeech      & Audiobooks                  & Close-talk mic.     & N              & CC-BY-4.0         \\
VoxPopuli        & European Parliament         & Close-talk mic.     & O              & CC0               \\
TED-LIUM         & TED talks                   & Close-talk mic.     & O              & CC-BY-NC-ND 3.0   \\
SwitchBoard      & Telephone conversations     & Telephone           & S              & LDC               \\
AMI              & Meetings                    & Headset             & S              & CC-BY-4.0         \\
\bottomrule
\end{tabular}
\end{center}
\end{table}

\noindent
\textbf{People's Speech} \citep{galvez21_peoples} is a large-scale English speech recognition dataset which assembles audio-transcription pairs sourced from the internet. The dataset covers multiple sources, including interviews, radio and finance. We use the ``clean" subset of the dataset, with approximately 12,000 hours of training data and corresponding validation and test splits.

\noindent
\textbf{GigaSpeech} \citep{chen21_gigaspeech} is a multi-domain speech recognition corpus curated from audiobooks, podcasts and YouTube. It contains both narrated and spontaneous speech over a range of content material, including science, arts and sports. We use the large subset (2,500 hours) to train and the standard validation and test splits.

\noindent
\textbf{Common Voice} \citep{ardila20_commonvoice} is a collection of open-license, crowd-source speech datasets where contributors record themselves narrating text from Wikipedia in various languages. Given its crowd-sourced approach, the dataset exhibits significant diversity in audio quality and speakers. The recorded audio often contains challenges such as background noise, accent variations, hesitations, and incorporation of non-native words. We use the English subset of version 13.0 (16-3-2023), with approximately 2,400 hours and the canonical data splits.

\noindent
\textbf{Fisher} \citep{cieri04_fisher} is a corpus of two-sided conversational telephone calls amongst speakers from the United States. We combine Part 1 \citep{ldc04_fisher_speech, ldc04_fisher_transcriptions} and Part 2 \citep{ldc04_fisher_speech_2, ldc04_fisher_transcriptions_2} of the dataset to give 1,960 hours of training data.

\noindent
\textbf{LibriSpeech} \citep{panayotov15_librispeech} is a standard dataset for training and evaluating academic speech models. It is comprised of 960 hours of narrated audiobooks sourced from the LibriVox\footnote{LibriVox: \url{https://librivox.org/}} project. The audiobook domain provides high-quality recording conditions, with little to no background noise. We use the standard split of train, validation (\emph{dev-clean}, \emph{dev-other}) and test sets (\emph{test-clean}, \emph{test-other}).

\noindent
\textbf{VoxPopuli} \citep{wang21_voxpopuli} is a large-scale multilingual speech datasets consisting of European Parliament event recordings from 2009-2020. The speech is oratory and from the political domain, with mostly non-native speakers. We use the English subset with approximately 550 hours and dataset splits provided therein.

\noindent
\textbf{TED-LIUM} \citep{hernandez18_tedlium} is a collection of English-language TED Talk conference videos. The talks span a variety of cultural, political, and academic themes. We use the Release 3 edition of the training set with approximately 450 hours and the legacy distribution of validation and test data.

\noindent
\textbf{SwitchBoard} \citep{godfrey92_switchboard} is a 260 hour corpus of two-sided conversational telephone calls amongst speakers from the United States. We partition 5\% of the SwitchBoard corpus to form the validation split The test sets are the Hub5Eval2000 \citep{ldc02_hub5} data with two subsets: SwitchBoard and CallHome.

\noindent
\textbf{AMI} \citep{carletta07_ami, renals07_ami} consists of 100 hours of meeting recordings captured using multiple recording streams in parallel. The corpus is manually annotated to provide the ground truth transcriptions. Individual samples of the AMI dataset contain very large audio files between 10 and 60 minutes in duration. We segment the audio samples according the the Kaldi \citep{Povey11_kaldi} recipe for AMI\footnote{AMI Kaldi recipe: \url{https://github.com/kaldi-asr/kaldi/tree/master/egs/ami/s5b}} to yield utterance of suitable length for training ASR systems. This involves splitting samples longer than 30 words at the time-stamps for punctuation to yield shorter utterances.

We use the individual headset microphone (AMI IHM) and single distant microphone (AMI SDM) versions of the dataset, with the train, validation and test sets provided therein.

\subsection{Short-Form Evaluation Data} \label{sec: appendix-short-form-data}

A detailed description of the short-form evaluation datasets is provided below.

\noindent
\textbf{CHiME-4} \citep{chime4_17} comprises of narrated samples from the Wall Street Journal corpus \citep{ldc93_wsj}. Recordings are performed in noisy environments using a 6-channel tablet based microphone array. We use the official 1-channel validation and test sets for evaluating our models.

\noindent
\textbf{Earnings-22} \citep{delrio22_earnings22} is a 119-hour test set of earnings calls recorded by global companies. The dataset was developed with the intention of assembling a diverse range of speakers and accents speaking in the context of real-world financial meetings. 

The Earnings-22 dataset contains audio recordings upwards of 10-minutes in duration. To create a short-form evaluation dataset, we segment these files into shorter samples up to 20-seconds in length. We first predict timestamps for the long audio files using the official \href{https://huggingface.co/patrickvonplaten/wav2vec2-base-960h-4-gram}{wav2vec 2.0 base + $4$-gram model} \citep{baevski20_wav2vec2} fine-tuned on the LibriSpeech \citep{baevski20_wav2vec2} dataset. We then split samples at the predicted timestamps for punctuation. If the samples are still longer than 20-seconds, we split them again at the longest silence in the utterance.

\noindent
\textbf{FLEURS} (Few-shot Learning Evaluation of Universal Representations of Speech) \citep{conneau22_fleurs} is a small-scale corpus for evaluating speech recognition systems in 102 languages. The transcription data is taken from the FLoRes-101 dataset \citep{goyal21_flores}, a machine translation corpus with 3001 samples of English text each translated to 101 other languages. To assemble the FLEURS dataset, up to three native speakers are recorded narrating the sentence translations in their native language. The recorded audio data is paired with the sentence transcriptions, thus yielding a multilingual speech recognition corpus. We use the English-US (en\_us) subset with 1 hour of validation and 2 hours of test data.

\noindent
\textbf{SPGISpeech} \citep{oneill21_spgispeech} is an English speech recognition dataset comprising of company earnings calls that have been manually transcribed by S\&P Global, Inc. We evaluate our models on the official validation and test splits, each of which is 100 hours.

\subsection{Long-Form Evaluation Data} \label{sec: appendix-long-form-data}

A detailed description of the long-form evaluation datasets is presented below.

\noindent
\textbf{Earnings-21} \citep{delrio21_earnings21} is a 39-hour corpus of company earnings calls over various financial sections.

\noindent
\textbf{Earnings-22} \citep{delrio22_earnings22} is a 119-hour test set of earnings calls recorded by global companies. The dataset was developed with the intention of assembling a diverse range of speakers and accents speaking in the context of real-world financial meetings. 

\noindent
\textbf{Meanwhile} \citep{radford22_whisper} is a collection of 64 segments taken from The Late Show with Stephen Colbert. The transcriptions are taken from the closed-caption data for each video and corrected with manual inspection. The YouTube URLs are provided by the Whisper authors, along with the segment start-end times\footnote{Meanwhile metadata: \url{https://github.com/openai/whisper/blob/main/data/meanwhile.json}}.

\noindent
\textbf{Rev 16} \citep{mohsin19_rev16} is a set of 30 podcast recordings that are commonly used to benchmark ASR systems in production settings. We follow the Whisper authors in evaluating on a subset of 16 of the 30 files with IDs:

$$
\texttt{3 4 9 10 11 14 17 18 20 21 23 24 26 27 29 32}
$$

\section{Experimental Set-Up}
\subsection{Training} \label{sec:appendix-training}

We train the models using the JAX and Flax neural network libraries \citep{jax18_github, heek20_flax}. We use data parallelism across TPU v4-8 accelerators \citep{jouppi20_tpu}, with bfloat16 precision and gradient checkpointing \citep{griewank00_checkpointing}. We use a softmax temperature of 2.0 to smooth the distributions of the student and teacher models when computing the KL divergence loss \citep{hinton15_knowledge}. Models are trained with an Optax implementation of the AdamW optimiser \citep{deepmind20_jax, loshchilov18_adamw} and gradient norm clipping \citep{pascanu13_gradclipping}. We train for a total of 80,000 optimisation steps, equivalent to eight epochs of training. We use the slanted triangular learning rate (STLR) \citep{howard18_universal} schedule, linearly increasing the learning rate from zero to a maximum of 1e-4 over the first 500 steps, and then linearly decaying it to zero. If the student encoder is the same as the teacher encoder, we freeze its parameters and only run it forward once. Back propagation is then only run through the distilled decoder. Table~\ref{tab: training hps} summarises the training hyperparemters.

\begin{table}[H]
\caption{Distil-Whisper training hyperparameters.}
\label{tab: training hps}
\begin{center}
\begin{tabular}{lc} 
\toprule
\textbf{Hyperparameter} & {\textbf{Value}}  \\ 
\midrule
Device                 & TPU v4-8                                 \\
Updates                 & 80,000                                 \\
Batch size              & 256                                 \\
Warmup steps            & 500                                 \\
LR schedule             & Linear decay                        \\
Precision               & bfloat16                            \\
KL softmax temperature           & 2.0                                 \\
Max grad norm           & 1.0                                 \\
Optimizer               & AdamW                               \\
$\beta_{1}$                   & 0.9                                 \\
$\beta_{2}$                   & 0.999                               \\
$\epsilon$                 & 10\textsuperscript{-8}                               \\
Weight decay            & 0.0                                 \\
Timestamp probability   & 50\%                                \\
\bottomrule
\end{tabular}
\end{center}
\end{table}

\subsection{Evaluation}

We evaluate all models in JAX on TPU v4-8 with greedy decoding unless specified otherwise. We normalise text using the Whisper English normaliser \citep{radford22_whisper}, which standardises text by removing or converting specific words, symbols, numeric expressions, and managing whitespace and spellings, in an attempt to only penalise a system when an error is caused by actually mistranscribing a word, and not by formatting differences. We measure transcription accuracy using the WER metric.

During training, we evaluate the intermediate checkpoints every 5k training steps on the 13 validation sets. We select the checkpoint with the best macro-average performance over the validation splits for final evaluation on the test splits.

For latency measurements, we evaluate the models in PyTorch \citep{paszke19_pytorch} using a single A100 40GB GPU in float16 precision. Specifically, we measure the total time taken to decode 256 samples from each of the four OOD test sets over batch sizes in the set \{1, 4, 16\}. Batch size 1 latency corresponds to short-form evaluation, where the models are evaluated without timestamp prediction. Batch sizes 4 and 16 correspond to long-form transcription, where the chunked long-form transcription algorithm is used. The Whisper models are evaluated with timestamp prediction and the Distil-Whisper models without. These configurations resulted in the best WER scores on the TED-LIUM long-form validation set.

Using the inference speed measurements, we compute the ratio of the inference time of Distil-Whisper to the Whisper large-v2 checkpoint, giving a figure for relative latency. We record all latency measurements using Flash Attention 2 \citep{dao23_flashattention2}, since it is a general inference optimisation for modern GPU hardware in production. In Section~\ref{sec: flash attention 2}, we show the effect of Flash Attention 2 on the latency of Whisper and Distil-Whisper.

\section{Evaluation Results} \label{sec:appendix-eval}
\begin{table}[H]
\caption{Per-dataset WER scores over the 15 short-form test sets. The macro-average WER scores are shown for the 11 ID datasets, four OOD datasets, and an overall average over all 15 test sets.}
\label{tab: full eval results}
\begin{center}
\begin{tabular}{l|rrrrr|rr} 
\toprule
\multicolumn{1}{l|}{\textbf{Dataset}} & \multicolumn{1}{l}{\textbf{\href{https://huggingface.co/openai/whisper-tiny.en}{tiny.en}}} & \multicolumn{1}{l}{\textbf{\href{https://huggingface.co/openai/whisper-base.en}{base.en}}} & \multicolumn{1}{l}{\textbf{\href{https://huggingface.co/openai/whisper-small.en}{small.en}}} & \multicolumn{1}{l}{\textbf{\href{https://huggingface.co/openai/whisper-medium.en}{medium.en}}} & \multicolumn{1}{l|}{\textbf{\href{https://huggingface.co/openai/whisper-large-v2}{large-v2}}} & \multicolumn{1}{l}{\textbf{\href{https://huggingface.co/distil-whisper/medium.en}{distil-medium.en}}} & \multicolumn{1}{l}{\textbf{\href{https://huggingface.co/distil-whisper/large-v2}{distil-large-v2}}}  \\ 
\midrule
AMI IHM                              & 22.9                              & 19.9                              & 17.4                               & 16.4                                & 16.9                                   & 16.1                                     & \textbf{14.7}                               \\
AMI SDM                              & 50.0                              & 45.2                              & 38.1                               & 37.0                                & 36.5                                   & 35.7                                     & \textbf{33.9}                               \\
Call Home                            & 23.8                              & 20.3                              & 19.0                               & 16.0                                & 17.5                                   & 15.1                                     & \textbf{13.5}                               \\
Common Voice 13                      & 28.9                              & 21.4                              & 15.3                               & 12.3                                & \textbf{10.4}                          & 15.3                                     & 12.9                                        \\
GigaSpeech                           & 13.5                              & 12.1                              & 11.0                               & 10.8                                & 10.7                                   & 11.2                                     & \textbf{10.5}                               \\
LibriSpeech clean                    & 5.9                               & 4.4                               & 3.3                                & 3.1                                 & \textbf{3.2}                           & 3.9                                      & 3.6                                         \\
LibriSpeech other                    & 14.1                              & 10.4                              & 7.4                                & 6.1                                 & \textbf{5.6}                           & 8.0                                      & 6.9                                         \\
People's Speech                      & 26.4                              & 22.2                              & 19.3                               & 18.6                                & 18.6                                   & 18.4                                     & \textbf{16.5}                               \\
SwitchBoard                          & 17.7                              & 15.6                              & 15.3                               & 14.0                                & 14.2                                   & 11.7                                     & \textbf{11.2}                               \\
TED-LIUM                             & 11.8                              & 10.9                              & 10.1                               & 11.5                                & 12.0                                   & 10.1                                     & \textbf{9.6}                                \\
Voxpopuli                            & 11.3                              & 9.6                               & 8.3                                & 7.9                                 & \textbf{7.3}                           & 8.8                                      & 8.0                                         \\ 
\midrule
CHIME-4                              & 32.7                              & 24.1                              & 15.7                               & 12.7                                & \textbf{11.8}                          & 15.1                                     & 14.0                                        \\
Earnings-22                          & 25.8                              & 21.2                              & 17.9                               & 17.0                                & \textbf{16.6}                          & 18.4                                     & 16.9                                        \\
FLEURS                               & 11.2                              & 7.5                               & 5.9                                & 4.9                                 & \textbf{4.2}                           & 6.9                                      & 6.3                                         \\
SPGISpeech                           & 5.8                               & 4.2                               & 3.6                                & 3.4                                 & 3.8                                    & 3.8                                      & \textbf{3.3}                                \\ 
\midrule
ID Average                           & 20.6                              & 17.5                              & 15.0                               & 14.0                                & 13.9                                   & 14.0                                     & \textbf{12.8}                               \\
OOD Average                          & 18.9                              & 14.3                              & 10.8                               & 9.5                                 & \textbf{9.1}                           & 11.1                                     & 10.1                                        \\
Average                              & 20.1                              & 16.6                              & 13.8                               & 12.8                                & 12.6                                   & 13.2                                     & \textbf{12.1}                               \\
\bottomrule
\end{tabular}
\end{center}
\end{table}

\begin{table}[H]
\caption{Per-dataset WER scores over the five long-form test sets. The macro-average WER scores are shown for the one ID dataset, four OOD datasets, and an overall average over all five test sets.}
\label{tab: long-form}
\begin{center}
\begin{tabular}{l|rrrrr|rr} 
\toprule
\multicolumn{1}{l|}{\textbf{Dataset}} & \multicolumn{1}{l}{\textbf{\href{https://huggingface.co/openai/whisper-tiny.en}{tiny.en}}} & \multicolumn{1}{l}{\textbf{\href{https://huggingface.co/openai/whisper-base.en}{base.en}}} & \multicolumn{1}{l}{\textbf{\href{https://huggingface.co/openai/whisper-small.en}{small.en}}} & \multicolumn{1}{l}{\textbf{\href{https://huggingface.co/openai/whisper-medium.en}{medium.en}}} & \multicolumn{1}{l|}{\textbf{\href{https://huggingface.co/openai/whisper-large-v2}{large-v2}}} & \multicolumn{1}{l}{\textbf{\href{https://huggingface.co/distil-whisper/medium.en}{distil-medium.en}}} & \multicolumn{1}{l}{\textbf{\href{https://huggingface.co/distil-whisper/large-v2}{distil-large-v2}}} \\ \midrule
TED-LIUM         & 6.4                           & 5.6                           & 5.8                           & 4.3                           & 4.4                            & 3.8                           & \textbf{3.7}                   \\ 
\midrule
Earnings 21      & 17.5                          & 14.5                          & 15.1                          & 12.3                          & 11.8                           & 11.6                          & \textbf{11.2}                  \\
Earnings 22      & 24.1                          & 19.4                          & 20.6                          & 15.6                          & \textbf{15.1}                  & 16.3                          & \textbf{15.1}                  \\
Meanwhile        & 16.4                          & 13.4                          & 8.7                           & 7.9                           & \textbf{6.3}                   & 8.9                           & 7.8                            \\
Rev 16           & 17.4                          & 15.4                          & 14.5                          & 13.2                          & 13.6                           & 13.0                          & \textbf{12.2}                  \\ 
\midrule
ID Average       & 6.4                           & 5.6                           & 5.8                           & 4.3                           & 4.4                            & 3.8                           & \textbf{3.7}                   \\
OOD Average      & 18.9                          & 15.7                          & 14.7                          & 12.3                          & 11.7                           & 12.4                          & \textbf{11.6}                  \\
Average          & 16.4                          & 13.7                          & 12.9                          & 10.7                          & 10.2                           & 10.7                          & \textbf{10.0}                  \\
\bottomrule
\end{tabular}
\end{center}
\end{table}

\begin{table}[H]
\caption{Per-dataset WER scores for the sequential and chunked long-form transcription algorithms. The macro-average WER scores are shown for the one ID dataset, four OOD datasets, and an overall average over all five test sets.}
\label{tab: algo full}
\begin{center}
\begin{tabular}{l|rr|r} 
\toprule
\multirow{2}{*}{\textbf{Dataset}} & \multicolumn{1}{l}{\textbf{\href{https://huggingface.co/openai/whisper-large-v2}{large-v2}}}   & \multicolumn{1}{l|}{\textbf{\href{https://huggingface.co/openai/whisper-large-v2}{large-v2}}} & \multicolumn{1}{l}{\textbf{\href{https://huggingface.co/distil-whisper/large-v2}{distil-large-v2}}}  \\
                                  & \multicolumn{1}{l}{\textbf{sequential}} & \multicolumn{1}{l|}{\textbf{chunked}}  & \multicolumn{1}{l}{\textbf{chunked}}          \\ 
\midrule
TED-LIUM                          & 4.0                                     & 4.4                                   & \textbf{3.7}                                           \\ 
\midrule
Earnings 21                       & \textbf{10.7}                                    & 11.8                                  & 11.2                                          \\
Earnings 22                       & \textbf{14.0}                                    & 15.1                                  & 15.1                                          \\
Meanwhile                         & \textbf{5.2}                                     & 6.3                                   & 7.8                                           \\
Rev 16                            & \textbf{11.7}                                    & 13.6                                  & 12.2                                          \\ 
\midrule
ID Average                        & 4.0                                     & 4.4                                   & \textbf{3.7}                                           \\
OOD Average                       & \textbf{10.4}                                    & 11.7                                  & 11.6                                          \\
Average                           & \textbf{9.1}                                     & 10.2                                  & 10.0                                          \\
\bottomrule
\end{tabular}
\end{center}
\end{table}

\begin{table}[H]
\caption{Per-dataset repeated 5-gram word duplicates (5-Dup.), insertion error rate (IER), substitution error rate (SER), deletion error rate (DER) and word error rate (WER) for the five long-form datasets. An average is shown for the ID dataset (TED-LIUM), the four OOD datasets, and an overall average.}
\label{tab: long-form hallucinations}
\begin{center}
\resizebox{\columnwidth}{!}{
\begin{tabular}{ll|r|rrrrr|rr} 
\toprule
\textbf{Dataset}                      & \textbf{Metric} & \multicolumn{1}{l|}{\textbf{\href{https://huggingface.co/facebook/wav2vec2-large-960h}{wav2vec2-large-960h}}} & \multicolumn{1}{l}{\textbf{\href{https://huggingface.co/openai/whisper-tiny.en}{tiny.en}}} & \multicolumn{1}{l}{\textbf{\href{https://huggingface.co/openai/whisper-base.en}{base.en}}} & \multicolumn{1}{l}{\textbf{\href{https://huggingface.co/openai/whisper-small.en}{small.en}}} & \multicolumn{1}{l}{\textbf{\href{https://huggingface.co/openai/whisper-medium.en}{medium.en}}} & \multicolumn{1}{l|}{\textbf{\href{https://huggingface.co/openai/whisper-large-v2}{large-v2}}} & \multicolumn{1}{l}{\textbf{\href{https://huggingface.co/distil-whisper/medium.en}{distil-medium.en}}} & \multicolumn{1}{l}{\textbf{\href{https://huggingface.co/distil-whisper/large-v2}{distil-large-v2}}}  \\ 
\midrule
\multirow{5}{*}{\textbf{TED-LIUM}}    & 5-Dup.          & \textbf{157}                  & 522                           & 557                           & 549                           & 452                           & 542                            & 283                           & 270                            \\
                                      & IER             & 1.7                           & 2.1                           & 2.1                           & 1.8                           & 1.4                           & 1.8                            & 0.6                           & \textbf{0.5}                   \\
                                      & SER             & 6.0                           & 2.2                           & 1.6                           & 1.2                           & 1.0                           & \textbf{0.9}                   & 1.4                           & 1.3                            \\
                                      & DER             & 1.9                           & 2.2                           & 1.9                           & 2.7                           & 2.0                           & \textbf{1.8}                   & \textbf{1.8}                  & \textbf{1.8}                   \\
                                      & WER             & 9.6                           & 6.4                           & 5.6                           & 5.8                           & 4.3                           & 4.4                            & 3.8                           & \textbf{3.7}                   \\ 
\midrule
\multirow{5}{*}{\textbf{Earnings-21}} & 5-Dup.          & \textbf{7938}                 & 19294                         & 19629                         & 20611                         & 21014                         & 21559                          & 16912                         & 16797                          \\
                                      & IER             & 5.2                           & 4.1                           & 3.2                           & 3.0                           & 3.0                           & 3.0                            & 2.0                           & \textbf{1.7}                   \\
                                      & SER             & 20.9                          & 8.2                           & 6.0                           & 4.6                           & 4.0                           & \textbf{3.9}                   & 5.0                           & 4.7                            \\
                                      & DER             & 4.3                           & 5.3                           & 5.3                           & 7.5                           & 5.3                           & 4.9                            & \textbf{4.5}                  & 4.7                            \\
                                      & WER             & 30.4                          & 17.5                          & 14.5                          & 15.1                          & 12.3                          & 11.8                           & 11.6                          & \textbf{11.2}                  \\ 
\midrule
\multirow{5}{*}{\textbf{Earnings-22}} & 5-Dup.          & \textbf{20869}                & 65599                         & 63041                         & 77122                         & 64977                         & 65419                          & 52475                         & 50949                          \\
                                      & IER             & 8.5                           & 6.5                           & 4.8                           & 5.3                           & 3.8                           & 3.9                            & 3.7                           & \textbf{3.0}                   \\
                                      & SER             & 26.8                          & 11.6                          & 8.5                           & 6.9                           & 5.9                           & \textbf{5.5}                   & 7.3                           & 6.7                            \\
                                      & DER             & 4.8                           & 6.0                           & 6.0                           & 8.4                           & 6.0                           & 5.7                            & \textbf{5.3}                  & 5.4                            \\
                                      & WER             & 40.1                          & 24.1                          & 19.4                          & 20.6                          & 15.6                          & \textbf{15.1}                  & 16.3                          & \textbf{15.1}                  \\ 
\midrule
\multirow{5}{*}{\textbf{Meanwhile}}   & 5-Dup.          & \textbf{858}                  & 1379                          & 1406                          & 1292                          & 1485                          & 1464                           & 1236                          & 1225                           \\
                                      & IER             & 1.5                           & 5.7                           & 5.2                           & 1.4                           & 3.6                           & 3.0                            & 1.4                           & \textbf{1.0}                   \\
                                      & SER             & 11.9                          & 9.1                           & 6.7                           & 4.2                           & 3.2                           & \textbf{2.4}                   & 5.5                           & 5.4                            \\
                                      & DER             & 3.0                           & 1.6                           & 1.5                           & 3.1                           & 1.0                           & \textbf{0.9}                   & 2.1                           & 1.4                            \\
                                      & WER             & 16.4                          & 16.4                          & 13.4                          & 8.7                           & 7.9                           & \textbf{6.3}                   & 8.9                           & 7.8                            \\ 
\midrule
\multirow{5}{*}{\textbf{Rev 16}}      & 5-Dup.          & \textbf{2220}                 & 6981                          & 6800                          & 6483                          & 6719                          & 6724                           & 5047                          & 5040                           \\
                                      & IER             & 4.2                           & 4.2                           & 3.8                           & 3.3                           & 3.4                           & 3.5                            & 2.8                           & \textbf{2.7}                   \\
                                      & SER             & 16.1                          & 6.8                           & 5.3                           & 4.2                           & 3.8                           & \textbf{3.7}                   & 4.6                           & 4.2                            \\
                                      & DER             & 6.1                           & 6.4                           & 6.3                           & 7.0                           & 6.0                           & 6.4                            & 5.6                           & \textbf{5.3}                   \\
                                      & WER             & 26.4                          & 17.4                          & 15.4                          & 14.5                          & 13.2                          & 13.6                           & 13.0                          & \textbf{12.2}                  \\ 
\midrule
\multirow{5}{*}{\textbf{ID Average}}  & 5-Dup.          & \textbf{157}                  & 587                           & 671                           & 548                           & 574                           & 752                            & 281                           & 270                            \\
                                      & IER             & 1.7                           & 2.4                           & 2.6                           & 1.9                           & 1.9                           & 2.7                            & 0.6                           & \textbf{0.5}                   \\
                                      & SER             & 6.0                           & 2.1                           & 1.6                           & 1.1                           & 1.0                           & \textbf{0.9}                   & 1.4                           & 1.3                            \\
                                      & DER             & 1.9                           & 2.2                           & 1.9                           & 2.0                           & 2.0                           & \textbf{1.7}                   & 1.8                           & 1.8                            \\
                                      & WER             & 9.6                           & 6.8                           & 6.0                           & 4.9                           & 4.8                           & 5.3                            & 3.8                           & \textbf{3.7}                   \\ 
\midrule
\multirow{5}{*}{\textbf{OOD Average}} & 5-Dup.          & \textbf{7971}                 & 23313                         & 22719                         & 26377                         & 23549                         & 23792                          & 18918                         & 18503                          \\
                                      & IER             & 4.8                           & 5.1                           & 4.3                           & 3.3                           & 3.5                           & 3.3                            & 2.5                           & \textbf{2.1}                   \\
                                      & SER             & 18.9                          & 8.9                           & 6.6                           & 5.0                           & 4.2                           & \textbf{3.9}                   & 5.6                           & 5.3                            \\
                                      & DER             & 4.6                           & 4.8                           & 4.8                           & 6.5                           & 4.6                           & 4.5                            & 4.4                           & \textbf{4.2}                   \\
                                      & WER             & 28.3                          & 18.9                          & 15.7                          & 14.7                          & 12.3                          & 11.7                           & 12.4                          & \textbf{11.6}                  \\ 
\midrule
\multirow{5}{*}{\textbf{Average}}     & 5-Dup.          & \textbf{6408}                 & 18755                         & 18287                         & 21211                         & 18929                         & 19142                          & 15191                         & 14856                          \\
                                      & IER             & 4.2                           & 4.5                           & 3.8                           & 3.0                           & 3.0                           & 3.0                            & 2.1                           & \textbf{1.8}                   \\
                                      & SER             & 16.4                          & 7.6                           & 5.6                           & 4.2                           & 3.6                           & \textbf{3.3}                   & 4.8                           & 4.5                            \\
                                      & DER             & 4.0                           & 4.3                           & 4.2                           & 5.7                           & 4.1                           & 4.0                            & 3.8                           & \textbf{3.7}                   \\
                                      & WER             & 24.6                          & 16.4                          & 13.7                          & 12.9                          & 10.7                          & 10.2                           & 10.7                          & \textbf{10.0}                  \\
\bottomrule
\end{tabular}}
\end{center}
\end{table}

\section{Additional Analysis}
\subsection{Early Exit}

Early exit is a paradigm for dynamically controlling the number of decoder layers used at inference time. It is based on the reasoning that the same amount of computation may not be required for every input to achieve adequate performance, depending on whether the input is easy or hard.

Instead of making a prediction based on the hidden-representation of the \textit{final} decoder layer, early exiting makes a prediction based on some \textit{intermediate} layer. For each decoder layer $l$, we compute a confidence score $c_{i}[l]$ for the $i$-th token. We also define an early-exit threshold $\alpha_{i}[l]$. If our confidence score exceeds this threshold ($c_{i}[l] > \alpha_{i}[l]$), we exit early and greedily predict the most probably token. Otherwise, we continue to the next layer and repeat.

Confident Adaptive Language Modeling (CALM) \citep{schuster22_calm} proposes using a softmax difference as the confidence score. The decoder hidden-state for the $l$-th layer $\boldsymbol{d}_{i}^{l}$ is mapped to the logit space using the word-embedding matrix $\boldsymbol{W}$. We then take the softmax of these logits to get the token probabilities from the $i$-th decoder layer: 

\begin{align}
    P\left( y_{i} | \boldsymbol{y}_{<i}, \boldsymbol{H}_{1:M}, \boldsymbol{d}_{i}^{l} \right) = \softmax \left(\boldsymbol{W} \boldsymbol{d}_{i}^{l}\right)
\end{align}

The confidence score $c_{i}[l]$ is defined as the difference between the top-2 most probable predictions. If this difference is greater than the threshold $\alpha_{i}[l]$, the model is confident of its predictions, and we can terminate decoding early.

To gauge how many decoder layers can be skipped with early exit, we benchmarked the performance of the Whisper medium.en model on 100 samples form the LibriSpeech test-clean dataset. As the dataset with the lowest WER performance on short-form evaluation (see Table~\ref{tab: full eval results}), it provides an upper-bound for the number of decoder layers that can be skipped, since the model should be most confident. We attempted setting the early-exit threshold automatically using the textual consistency formulation from CALM, which guarantees that the model will perform to within a certain tolerance of the full model with specified probability, but found it skipped close to zero layers for almost all examples. Instead, we swept over a set of values of the threshold, recording the WER performance and number of decoder layers used.

Table~\ref{tab: calm} shows the average number of decoder layers utilised by the medium.en model as the early-exit threshold is reduced. The medium.en model has a total of 24 decoder layers, of which the last 3 are skipped almost immediately. However, the WER penalty is significant even for just a 3-layer reduction. As we reduce the threshold, the number of layers skipped does not reduce significantly, but the WER penalty continues to inflate. Setting the threshold to 0.9750 results in an average of 3 skipped decoder layers, yielding an inference speed-up of 1.1 times. However, it also causes an increase in WER from 2.3\% to 3.4\% This suggests that there is high-utilisation of the first 21 decoder layers in the pre-trained Whisper model, and that the final 3 layers are necessary for ensuring high transcription accuracy. We leave finding effective early exit schemes for Seq2Seq ASR models as future work.

\begin{table}
\caption{\textbf{Early-exit performance.} WER on 100 examples from the LibriSpeech test-clean dataset as the early-exit threshold is varied. The latency results are computed relative to the medium.en model with full utilisation of the 24 decoder layers.}
\label{tab: calm}
\begin{center}
\begin{tabular}{rrrr} 
\toprule
\multicolumn{1}{l}{\textbf{Threshold}} & \multicolumn{1}{l}{\textbf{Avg. Dec Layers}} & \multicolumn{1}{l}{\textbf{Rel. Latency}} & \multicolumn{1}{l}{\textbf{WER}}  \\ 
\midrule
1.0000                                 & 24.0                                             & 1.0                                       & 2.3                               \\
0.9875                                 & 21.2                                             & 1.1                                       & 2.8                               \\
0.9750                                 & 21.0                                             & 1.1                                       & 3.4                               \\
0.9625                                 & 20.8                                             & 1.1                                       & 3.5                               \\
0.9500                                 & 20.7                                             & 1.2                                       & 3.6                               \\
0.9375                                 & 20.6                                             & 1.2                                       & 3.7                               \\
0.9250                                 & 20.5                                             & 1.2                                       & 4.3                               \\
\bottomrule
\end{tabular}
\end{center}
\end{table}

\subsection{Distillation Objective}

The knowledge distillation (KD) objective proposed in Section~\ref{sec: knowledge distillation} is a weighted average of the Kullback-Leibler (KL) divergence and pseudo-label (PL) terms:

\begin{align}
    \mathcal{L}_{KD} = \alpha_{KL}  \mathcal{L}_{KL} + \alpha_{PL} \mathcal{L}_{PL}
\end{align}

The typical setting in layer-based compression is that the dimensionality of the student model matches that of the teacher model. This means the student and teacher layers output the same shape of hidden-states. Thus, we can introduce a mean-square error (MSE) term to encourage the student's hidden layer outputs to match those of the teacher:

\begin{align}
    \mathcal{L}_{MSE} = \sum_{i=1}^{N}\sum_{l=1}^{L'} MSE \left( \boldsymbol{H}_{l}^{S}, \boldsymbol{H}_{\phi\left(l \right)}^{T}\right)
\end{align}

where $\boldsymbol{H}_{l}^{S}$ is the hidden-state output from the $l$-th layer of the student model $S$, $\phi\left(l\right)$ maps the $l$-th student layer to the corresponding teacher layer it is trained to emulate, and $\boldsymbol{H}_{\phi\left(l\right)}^{T}$ is the hidden-state output from layer $\phi\left(l\right)$ of the teacher model $T$. The mapping $\phi$ follows the settings from \cite{schleifer20_distilbart}, where it is selected such that each decoder layer is trained to behave like maximally spaced teacher layers. For example, given a 2-layer student model initialised from a 32-layer teacher model, we choose pairings in $\phi$ such that each student decoder layer is taught to behave like 16 teacher layers. Thus, student layer 1's hidden-states are paired to teacher layer 16, and student layer 2's hidden-states paired to teacher layer 32:

\begin{equation}
    \phi(l) = \begin{cases} 16 & \text{if } l = 1 \\ 32 & \text{if } l = 2 \end{cases}
\end{equation}

A more general KD training objective is then a weighted sum of the KL, PL and MSE terms:

\begin{align}
    \mathcal{L}_{KD} = \alpha_{KL} \mathcal{L}_{KL} + \alpha_{PL} \mathcal{L}_{PL} + \alpha_{MSE} \mathcal{L}_{MSE}
\end{align}

where $\alpha_{KL}$, $\alpha_{PL}$ and $\alpha_{MSE}$ are scalar weights for the KL, PL and MSE loss terms respectively. Following \citep{schleifer20_distilbart}, we set $\alpha_{KL} = 0.8$ and $\alpha_{PL} = 1.0$, and tune the value of $\alpha_{MSE}$ on our validation set.

To quantify the performance gain obtained by incorporating each KD term, we train distil-large-v2 checkpoints for 10,000 training steps (two epochs) on a three combinations of KD objectives: (i) PL, (ii) PL + KD, and (iii) PL + KD + MSE. Training on PL alone is equivalent to shrink and fine-tune (SFT), but with the ground truth labels replaced by the PL generated ones. In all cases, Whisper-generated pseudo-labels are used as the ground truth labels during training. 

Table~\ref{tab: training objective} displays the average WER across the 11 short-form in-distribution (ID) validation sets and the three out-of-distribution (OOD) validation sets for each KD combination. The addition of the KL-divergence term yields an OOD word error rate (WER) that is 0.3\% absolute lower compared to just PL. This suggests that the additional information transferred from the teacher to the student during KD is beneficial over training on PL alone. Incorporating the MSE loss term had a negligible effect on the average WER performance of the distilled model. This indicates that there is sufficient training signal from the PL and KL loss terms. The MSE loss requires that the hidden-states for each layer are recorded and kept in memory. This added a significant overhead when training the model in JAX, which resulted in a decrease to the maximum possible batch size. By only using the PL and KL objectives and training with a higher throughput, we achieved better results within a specified time interval compared to using the MSE loss, and thus opted for this configuration for our experiments. Therefore, our final KD training objective is a weighted sum of the PL and KL terms only.

\begin{table}
\caption{\textbf{Impact of the distillation objective.} Average WER of the distil-large-v2 checkpoint over the 11 ID and three OOD validation sets for the three possible training objectives: pseudo-labels (PL), KL-divergence (KL) and mean-square error (MSE).}
\label{tab: training objective}
\begin{center}
\begin{tabular}{lrrr} 
\toprule
\textbf{Objective} & \multicolumn{1}{l}{\textbf{Avg. ID WER}} & \multicolumn{1}{l}{\textbf{Avg. OOD WER}} & \multicolumn{1}{l}{\textbf{Avg. WER}}  \\ 
\midrule
PL                 & 12.8                                     & 7.7                                       & 11.6                                   \\
PL + KL            & 12.6                                     & 7.4                                       & 11.4                                   \\
PL + KL + MSE      & 12.6                                     & 7.3                                       & 11.4                                   \\
\bottomrule
\end{tabular}
\end{center}
\end{table}

\subsection{Batched Speculative Decoding}

Table~\ref{tab: batched spec dec} reports the relative latency of the medium.en and large-v2 models both with and without speculative decoding at various batch sizes. For a batch size of 1, speculative decoding with the distil-large-v2 assistant yields a 2.0 times increase to inference speed over the large-v2 alone. This speed-up is comparable to using the tiny model as the assistant. For the medium.en model, using distil-medium.en as the assistant provides a 2.4 times speed-up. This outperforms the tiny.en assistant, which is only 2.0 times faster. A similar trend holds for a batch size of 4, albeit with smaller improvements to relative latency. For a batch size of 16, speculative decoding performs worse than using the main model by itself. For batched speculative decoding, all candidate tokens in across the batch must match the validation tokens for the tokens to be accepted. If any token in the batch at a given position does not agree, all candidate tokens that precede the position are discarded. Consequently, speculative decoding favours lower batch sizes, where it provides significant latency improvements while ensuring the same outputs as the original model.

\begin{table}
\caption{\textbf{Impact of speculative decoding with batching.} Relative latency of the medium.en and large-v2 models using either the tiny Whisper or Distil-Whisper assistant models. The assistant models used are shown below the main ones. The relative latency is computed relative to the large-v2 model without speculative decoding for each batch size.}
\label{tab: batched spec dec}
\begin{center}
\begin{tabular}{l|r|rrr} 
\toprule
\multirow{2}{*}{\textbf{Model}} & \multicolumn{1}{l|}{\multirow{2}{*}{\textbf{Params / M}}} & \multicolumn{3}{c}{\textbf{Batch Size}}                                 \\
                                & \multicolumn{1}{l|}{}                                     & \multicolumn{1}{c}{1} & \multicolumn{1}{c}{4} & \multicolumn{1}{c}{16}  \\ 
\midrule
\href{https://huggingface.co/openai/whisper-medium.en}{medium.en}                      & 769                                                       & 1.4                   & 1.3                   & 1.5                     \\
w\textbackslash{} \href{https://huggingface.co/openai/whisper-tiny.en}{tiny.en}                       & 808                                                       & 2.7                   & 1.8                   & 1.2                     \\
w\textbackslash{} \href{https://huggingface.co/distil-whisper/medium.en}{distil-medium.en}              & 856                                                       & 3.3                   & 2.2                   & 1.3                     \\ 
\midrule
\href{https://huggingface.co/openai/whisper-large-v2}{large-v2}                        & 1550                                                      & 1.0                   & 1.0                   & 1.0                     \\
w\textbackslash{} \href{https://huggingface.co/openai/whisper-tiny}{tiny}                          & 1589                                                      & 2.1                   & 1.3                   & 0.8                     \\
w\textbackslash{} \href{https://huggingface.co/distil-whisper/large-v2}{distil-large-v2}               & 1669                                                      & 2.0                   & 1.3                   & 0.8                     \\
\bottomrule
\end{tabular}
\end{center}
\end{table}

\subsection{Strategies for Reliable Long-form Transcription} \label{sec: chunk analysis}

Transcribing long-form audio relies on the accurate prediction of multiple chunks of audio in parallel. Since long-form audio typically contains instances of long pauses between spoken utterances, the Whisper model has a higher propensity to hallucinate compared to short-form audio. To combat this, the hyper-parameters of the chunked long-form transcription algorithm can be optimised to help avoid failure cases. These tuned hyper-parameters are applied in the long-form transcription results reported in Section~\ref{sec: long-form eval}.

Table~\ref{tab: chunk length} shows the WER performance on the long-form TED-LIUM validation set as the chunk length of the audio segments is decreased for the large-v2 and distil-large-v2 models. For Whisper, a chunk length of 30-seconds is optimal, whereas Distil-Whisper performs best with a chunk length of 15-seconds, giving the  best overall performance on the validation set with 4.1\% WER. Whisper is pre-trained on 30-seconds audio samples, whereas the mean sample length in the Distil-Whisper training set is 7.1-seconds (see Appendix~\ref{sec: appendix-training-data} for details). This suggests that the chunk length should be selected based on the distribution of audio data the model is trained on.

It is worth noting that the long-form WERs of Whisper and Distil-Whisper on the TED-LIUM validation set are 2.4\% and 3.0\% WER lower than their short-form performance on the same dataset. This performance gain demonstrates that the long-form algorithm used by Distil-Whisper is an effective approach for transcribing long audio files with strong WER performance.

\begin{table}
\caption{\textbf{Effect of chunk length on the chunked long-form algorithm.} WER performance on the long-form TED-LIUM validation set as the chunk length of the long-form transcription algorithm is reduced.}
\label{tab: chunk length}
\begin{center}
\begin{tabular}{rrr} 
\toprule
\multicolumn{1}{l}{\textbf{Chunk Length / s}} & \multicolumn{1}{l}{\textbf{large-v2}} & \multicolumn{1}{l}{\textbf{distil-large-v2}}  \\ \midrule
30                                        & 4.8         & 7.4                                                                        \\
25                                        & 5.3         & 5.7                                                                         \\
20                                        & 6.5         & 5.0                                                                         \\
15                                        & 6.5         & 4.2                                                                         \\
10                                        & 10.0        & 4.3                                                                         \\
\bottomrule
\end{tabular}
\end{center}
\end{table}

\subsection{Flash Attention 2} \label{sec: flash attention 2}


Flash Attention (FA) \citep{dao22_flashattention} addresses the slow and memory-intensive nature of transformers models on long sequences by proposing an IO-aware exact attention algorithm. It uses tiling to minimise memory reads/writes between GPU high bandwidth memory (HBM) and GPU on-chip memory (SRAM), resulting in faster inference and the ability to handle longer sequence lengths with improved model performance. Flash Attention 2 (FA2) \citep{dao23_flashattention2} further refines this approach, optimising GPU work partitioning to achieve even greater computational efficiency. 

To demonstrate the effect of FA2 on the Whisper and Distil-Whisper models, we benchmark the inference speed over 256 examples from each of the four OOD test sets. We report the latency as the real-time factor (RTF), defined as the ratio of the inference time to the audio duration. In order to generalise to multiple hardware, we report the inference time on a 40GB A100 GPU, which is typically used in industry, as well as a 16GB T4 GPU, a typical consumer-grade GPU. The results serve as look-up tables for practitioners wishing to determine the best trade-off between WER performance and latency for various batch sizes and hardware.

Table~\ref{tab: inference optimisations a100} reports the RTF for batch sizes 1, 4 and 16 on a 40GB A100 GPU. For the base attention implementation, distil-large-v2 is faster than base.en at batch sizes 1 and 4, and marginally slower at batch size 16. It remains faster than small.en at batch size 16, with an average OOD WER that is 0.7\% lower. For all batch sizes, distil-large-v2 is at least 3.3 times faster than large-v2, while performing to within 1\% WER. This highlights that distil-large-v2 can be used as a drop-in replacement for base.en at low batch sizes and small.en at higher ones, with 3.3\% and 0.7\% WER improvements respectively. Similarly, distil-medium.en is faster than tiny.en at batch sizes 1 and 4, and faster than base.en at batch size 16, while performing 3.2\% better on OOD test data.

Incorporating FA2 benefits Distil-Whisper more than Whisper at higher batch sizes: the distil-large-v2 model is 31\% faster with FA2 at batch size 16, compared to 22\% for large-v2. Furthermore, the distil-medium.en checkpoint is as fast as tiny.en at batch size 16, with an average OOD WER that is 7.8\% lower. FA2 has a greater improvement on the encoder than the decoder, where the memory is re-allocated at each decoding step. Since Distil-Whisper consists of 32 encoder layers and only 2 decoder layers, it improves the inference time significantly more than Whisper, which has 32 encoder and 32 decoder layers. This suggests that FA2 should always be incorporated for Distil-Whisper when operating at higher batch sizes. Using a static key/value cache would result in a more significant speed-up to the inference time of the decoder. We leave this as future works.

\begin{table}
\caption{Real time factor (RFT) with and without FA2 for batch sizes 1, 4 and 16. Inference speed is measured on a 40GB A100 GPU with PyTorch 2.0. RTF is expressed in 10\textsuperscript{-3}.}
\label{tab: inference optimisations a100}
\begin{center}
\begin{tabular}{l|r|rrr|rrr} 
\toprule
\multirow{2}{*}{\textbf{Model}} & \multicolumn{1}{l|}{\multirow{2}{*}{\textbf{Avg. OOD WER}}} & \multicolumn{3}{c|}{\textbf{Base}} & \multicolumn{3}{c}{\textbf{Flash Attention 2}}  \\
                                & \multicolumn{1}{l|}{}                                       & 1     & 4    & 16                  & 1     & 4    & 16                               \\ 
\midrule
\href{https://huggingface.co/openai/whisper-tiny.en}{tiny.en}                         & 18.9                                                        & 22.7  & 8.7  & \textbf{3.1}                 & 21.4  & 8.3  & \textbf{2.9}                              \\
\href{https://huggingface.co/openai/whisper-base.en}{base.en}                         & 14.3                                                        & 27.9  & 11.7 & 4.6                 & 30.5  & 10.7 & 3.4                              \\
\href{https://huggingface.co/openai/whisper-small.en}{small.en}                        & 10.8                                                        & 59.1  & 22.3 & 8.4                 & 50.7  & 18.3 & 6.3                              \\
\href{https://huggingface.co/openai/whisper-medium.en}{medium.en}                       & 9.5                                                         & 99.4  & 43.1 & 15.1                & 89.3  & 37.1 & 11.2                             \\
\href{https://huggingface.co/openai/whisper-large-v2}{large-v2}                        & \textbf{9.1}                                                         & 137.2 & 54.9 & 21.3                & 121.9 & 48.4 & 16.6                             \\ 
\midrule
\href{https://huggingface.co/distil-whisper/medium.en}{distil-medium.en}                & 11.1                                                        & \textbf{19.2}  & \textbf{7.5}  & 4.3                 & \textbf{20.4}  & \textbf{7.1}  & \textbf{2.9}                              \\
\href{https://huggingface.co/distil-whisper/large-v2}{distil-large-v2}                 & 10.1                                                        & 25.1  & 9.9  & 6.5                 & 24.2  & 8.9  & 4.4                              \\
\bottomrule
\end{tabular}
\end{center}
\end{table}

Table~\ref{tab: inference optimisations t4} reports the RTF on a 16GB T4 GPU. Since FA2 is not currently supported on T4, we report the RTF using the original FA implementation. For the base attention implementation, distil-large-v2 is faster than small.en at batch size 1. However, the distilled models follow the trend of medium.en and large-v2 at higher batch sizes, where the percentage decreases to RTF are much lower than those of the smaller pre-trained Whisper checkpoints, such as base.en and small.en. At batch size 4 and 16, distil-large-v2 is slower than small.en. The distil-medium.en model is faster than small.en at batch size 4, but slower at 16. 

This performance pattern can be attributed to the Distil-Whisper architecture: since we copy the entire encoder from the original Whisper medium.en and large-v2 models, the distil-medium.en and distil-large-v2 models require more memory than small.en, especially at higher batch sizes. As the memory on a T4 GPU increases, the throughput saturates, resulting in diminishing RTF benefits at batch sizes of 4 and 16. This trend is also observed for the medium.en and large-v2 Whisper checkpoints.

The latency improvement obtained using FA is significant for both Whisper and Distil-Whisper. At batch size 1, distil-large-v2 is comparable to base.en, while distil-medium.en is faster than tiny.en. However, the memory savings are not enough to offset the effects of the T4 GPU at higher batch sizes; distil-large-v2 is slower than small.en at batch size 4 and 16, and distil-medium.en slower than base.en.

Overall, a T4 GPU may be adequate for operating Whisper and Distil-Whisper models at a batch size of 1. For batch sizes beyond this, there is a notable performance stagnation on a T4, and higher memory A100 GPUs are preferential.

\begin{table}
\caption{Real time factor (RFT) with and without FA for batch sizes 1, 4 and 16. Inference speed is measured on a 16GB T4 GPU with PyTorch 2.0. RTF is expressed in 10\textsuperscript{-3}.}
\label{tab: inference optimisations t4}
\begin{center}
\begin{tabular}{l|r|rrr|rrr} 
\toprule
\multirow{2}{*}{\textbf{Model}} & \multicolumn{1}{l|}{\multirow{2}{*}{\textbf{Avg. OOD WER}}} & \multicolumn{3}{c|}{\textbf{Base}} & \multicolumn{3}{c}{\textbf{Flash Attention}}  \\
                                & \multicolumn{1}{l|}{}                                       & 1     & 4    & 16                  & 1     & 4    & 16                               \\ 
\midrule
\href{https://huggingface.co/openai/whisper-tiny.en}{tiny.en}                           & 18.9                                                        & \textbf{20.0}  & \textbf{7.2}  & \textbf{2.6}                 & 21.2  & \textbf{6.9}  & \textbf{2.5}                              \\
\href{https://huggingface.co/openai/whisper-base.en}{base.en}                           & 14.3                                                        & 26.1  & 9.5  & 4.2                 & 26.1  & 9.6  & 3.7                              \\
\href{https://huggingface.co/openai/whisper-small.en}{small.en}                         & 10.8                                                        & 48.3  & 19.0 & 9.9                 & 42.6  & 16.8 & 8.0                              \\
\href{https://huggingface.co/openai/whisper-medium.en}{medium.en}                       & 9.5                                                         & 89.9  & 44.6 & 26.5                & 66.2  & 33.8 & 18.5                             \\
\href{https://huggingface.co/openai/whisper-large-v2}{large-v2}                         & \textbf{9.1}                                                         & 129.0 & 74.5 & 47.5                & 100.6 & 52.0 & 33.8                             \\ 
\midrule
\href{https://huggingface.co/distil-whisper/medium.en}{distil-medium.en}                & 11.1                                                        & 23.0  & 18.1 & 17.2                & \textbf{19.0}  & 12.1 & 10.4                             \\
\href{https://huggingface.co/distil-whisper/large-v2}{distil-large-v2}                  & 10.1                                                        & 38.0  & 31.8 & 31.2                & 27.4  & 20.8 & 20.1                             \\
\bottomrule
\end{tabular}
\end{center}
\end{table}

\end{document}